\documentclass[letterpaper]{article} % DO NOT CHANGE THIS
\usepackage[preprint]{aaai2027}  % DO NOT CHANGE THIS
\usepackage[hyphens]{url}  % DO NOT CHANGE THIS
\usepackage{graphicx} % DO NOT CHANGE THIS
\usepackage{natbib}  % DO NOT CHANGE THIS AND DO NOT ADD ANY OPTIONS TO IT
\usepackage{caption} % DO NOT CHANGE THIS AND DO NOT ADD ANY OPTIONS TO IT
\usepackage{algorithm}
\usepackage{algpseudocode}  % 自己加的
\algrenewcommand\algorithmicrequire{\textbf{Input:}}
\algrenewcommand\algorithmicensure{\textbf{Output:}}
\algnewcommand{\algorithmicauxiliary}{\textbf{Auxiliary:}}
\algnewcommand{\Auxiliary}{\item[\algorithmicauxiliary]}

\usepackage[table]{xcolor}
\definecolor{rowblue}{RGB}{235,242,252} % 浅蓝色
\usepackage{newfloat}
\usepackage{listings}
\DeclareCaptionStyle{ruled}{labelfont=normalfont,labelsep=colon,strut=off} % DO NOT CHANGE THIS
\floatstyle{ruled}
\newfloat{listing}{tb}{lst}{}
\floatname{listing}{Listing}

\usepackage{booktabs}

\usepackage{amsmath}
\usepackage{makecell}

\usepackage{tcolorbox}
\definecolor{boxback}{RGB}{245, 245, 245} % 浅灰背景
\definecolor{boxframe}{RGB}{100, 100, 100} % 深灰边框

\usepackage{tcolorbox}
\tcbuselibrary{skins, breakable}
\usepackage{booktabs} % 用于专业表格线
\usepackage{pifont}   % 用于对勾叉号
\usepackage{enumitem} % 用于紧凑列表

\newtcolorbox{casestudybox}[2][]{%
    enhanced,
    colframe=gray!60!black,   % 边框颜色（深灰）
    colbacktitle=gray!70!black, % 标题背景色
    coltitle=white,           % 标题文字颜色
    colback=white,            % 内容背景色
    fonttitle=\bfseries\large,
    arc=1mm, boxrule=0.5mm,   % 圆角和边框粗细
    title={#2},               % 标题内容
    #1                        % 可选参数（如 label）
}

\usepackage{tcolorbox}
\usepackage{xcolor}
\usepackage{tikz}
\usepackage{pifont}
\usepackage{enumitem}
\usetikzlibrary{arrows.meta}

\definecolor{tasnavy}{RGB}{5,35,80}
\definecolor{tasred}{RGB}{180,25,25}
\definecolor{tasblue}{RGB}{20,85,180}
\definecolor{tasorange}{RGB}{170,105,0}
\definecolor{tasgreen}{RGB}{45,130,35}
\definecolor{taspurple}{RGB}{120,55,190}
\definecolor{tasgray}{RGB}{245,245,245}

\newcommand{\cmark}{\textcolor{green!60!black}{\ding{51}}}

\newcommand{\turnarrow}[1]{%
\tikz[baseline=-3.9ex]{
  \draw[#1, semithick, rounded corners=1pt, -{Stealth[length=1.7mm]}]
  (0,0.22) -- (0,-0.08) -- (0.45,-0.08);
}%
}

\newcommand{\treenode}[5]{%
\par\vspace{0.28em}
\noindent
\hspace*{#1}
\turnarrow{#2}
\hspace{0.35em}
\begin{tcolorbox}[
    enhanced,
    colback=#2!4,
    colframe=#2!75!black,
    boxrule=0.55pt,
    arc=1.2mm,
    left=1.5mm,
    right=1.5mm,
    top=1mm,
    bottom=1mm,
    width=#3,
    boxsep=0.6mm,
    nobeforeafter
]
\textbf{\textcolor{#2!85!black}{#4}}\\[-0.1em]
{\footnotesize #5}
\end{tcolorbox}
\par
}

\newcommand{\rootnode}[3]{%
\par\vspace{0.25em}
\noindent
\begin{tcolorbox}[
    enhanced,
    colback=#1!4,
    colframe=#1!80!black,
    boxrule=0.65pt,
    arc=1.2mm,
    left=1.8mm,
    right=1.8mm,
    top=1mm,
    bottom=1mm,
    width=#2,
    boxsep=0.7mm,
    nobeforeafter
]
#3
\end{tcolorbox}
\par
}

\newcommand{\elbowarrow}[1]{%
\tikz[baseline=-2.0ex]{
  \draw[#1, semithick, rounded corners=1pt, -{Stealth[length=1.5mm]}]
  (0,0.16) -- (0,-0.10) -- (0.38,-0.10);
}%
}
\newcommand{\compactelbowarrow}[1]{%
\tikz[baseline=-0.45ex]{
  \draw[#1, semithick, rounded corners=1pt, -{Stealth[length=1.5mm]}]
  (0,0.16) -- (0,-0.10) -- (0.38,-0.10);
}%
}

\newcommand{\compactroot}[4]{%
\par\vspace{0.16em}
\noindent
\begin{tcolorbox}[
    enhanced,
    colback=#1!4,
    colframe=#1!80!black,
    boxrule=0.55pt,
    arc=1mm,
    left=1.2mm,
    right=1.2mm,
    top=0.8mm,
    bottom=0.8mm,
    width=#2,
    boxsep=0.5mm,
    nobeforeafter
]
{\small\textbf{\textcolor{#1!85!black}{#3}} #4}
\end{tcolorbox}
\par
}

\newcommand{\compactnode}[5]{%
\par\vspace{0.16em}
\noindent
\hspace*{#1}
\elbowarrow{#2}
\hspace{0.28em}
\begin{tcolorbox}[
    enhanced,
    colback=#2!4,
    colframe=#2!75!black,
    boxrule=0.5pt,
    arc=1mm,
    left=1.2mm,
    right=1.2mm,
    top=0.75mm,
    bottom=0.75mm,
    width=#3,
    boxsep=0.5mm,
    nobeforeafter
]
{\small\textbf{\textcolor{#2!85!black}{#4}}: #5}
\end{tcolorbox}
\par
}

\newcommand{\atomagent}[2]{%
\par\vspace{0.05em}
\hspace*{1.15em}
\compactelbowarrow{tasorange}
\hspace{0.15em}
{\scriptsize\textbf{atom #1:} collect attributes for \textit{#2}.}
}

\newcommand{\exptransfer}[1]{%
\par\vspace{0.03em}
\hspace*{1.0em}
\compactelbowarrow{taspurple}
\hspace{0.05em}
{\scriptsize\textbf{\textcolor{taspurple!85!black}{experience:}}
#1
}
}

\newcommand{\brandwideexpanded}[5]{%
\begin{tcolorbox}[
    enhanced,
    colback=tasblue!3,
    colframe=tasblue!65!black,
    boxrule=0.45pt,
    arc=1mm,
    left=1.1mm,
    right=1.1mm,
    top=0.7mm,
    bottom=0.7mm,
    width=\linewidth,
    boxsep=0.45mm,
    nobeforeafter
]
{\footnotesize
\textbf{\textcolor{tasblue!85!black}{wide(depth=1, #1)}} 
\hfill 
\textcolor{tasgreen!70!black}{parallel branch #2/6}

\vspace{0.12em}
\compactelbowarrow{gray}
{\scriptsize\textbf{entity\_collect:}
 collect the core/permanent product list for #1.}\\
\vspace{0.08em}
\hspace*{1.15em}
{\scriptsize\textbf{Output:} #5}

\vspace{0.12em}
\compactelbowarrow{tasorange}
{\scriptsize\textbf{assign #3 atom agents:}}
#4
}
\end{tcolorbox}
\vspace{0.25em}
}

\newcommand{\dwvehiclewideexpanded}[5]{%
\begin{tcolorbox}[
    enhanced,
    colback=tasblue!3,
    colframe=tasblue!65!black,
    boxrule=0.45pt,
    arc=1mm,
    left=1.1mm,
    right=1.1mm,
    top=0.7mm,
    bottom=0.7mm,
    width=\linewidth,
    boxsep=0.45mm,
    nobeforeafter
]
{\footnotesize
\vspace{0.12em}
\compactelbowarrow{gray}
\textbf{entity\_collect:}
find all #1 first launched or resumed US production from 2010--2024,
excluding facelifts/model updates.\\
\vspace{0.18em}
\hspace*{1.15em}
{\textbf{Output:} #5}

\vspace{0.22em}
\compactelbowarrow{tasorange}
\textbf{assign #3 atom agents:}

\vspace{-0.50em}
#4
}
\end{tcolorbox}
\vspace{0.25em}
}

\newcommand{\explorecontext}[1]{%
\begin{tcolorbox}[
    enhanced,
    colback=taspurple!3,
    colframe=taspurple!70!black,
    boxrule=0.5pt,
    arc=1mm,
    left=1.2mm,
    right=1.2mm,
    top=0.8mm,
    bottom=0.8mm,
    width=#1,
    boxsep=0.5mm,
    nobeforeafter
]
{\small
\textbf{\textcolor{taspurple!85!black}{web-probing agent: web structure hint}}\\[-0.15em]
\textbf{Coverage:} comprehensive. 
\textbf{Sources:} Wikipedia tour pages, Concert Archives, and tour-level live-performance summaries.\\
\textbf{Data organization:} sources are either tour-based or reverse-chronological. 
The efficient strategy is to split by official tour, extract show-level tables, filter dates, then merge chronologically.
}
\end{tcolorbox}
}

\newcommand{\atomtour}[3]{%
\begin{tcolorbox}[
    enhanced,
    colback=tasorange!4,
    colframe=tasorange!75!black,
    boxrule=0.45pt,
    arc=1mm,
    left=1.1mm,
    right=1.1mm,
    top=0.7mm,
    bottom=0.7mm,
    width=\linewidth,
    boxsep=0.45mm,
    nobeforeafter
]
{\footnotesize
\textbf{\textcolor{tasorange!90!black}{atom #2: #1}}
\hfill
\textcolor{tasgreen!70!black}{parallel}

\vspace{0.12em}
\textbf{source:} dedicated Wikipedia tour page, \texttt{#3}

\vspace{0.12em}
\textbf{task:} extract every official show within Jan. 1, 2010--May 1, 2025.
}
\end{tcolorbox}
\vspace{0.22em}
}

\newcommand{\dwvehicleatom}[2]{%
\par\vspace{0.05em}
\hspace*{1.1em}
\compactelbowarrow{tasorange}
\hspace{0.12em}
{\scriptsize\textbf{atom #1:} collect attributes for \textit{#2}.}
}

\definecolor{PromptBack}{RGB}{245,246,255}
\tcbuselibrary{listings,skins}
\usepackage{caption}
\usepackage{upquote}
\definecolor{PromptBlack}{RGB}{0,0,0}

\newtcblisting{promptbox}[1]{
  enhanced,
  listing only,
  listing engine=listings,
  width=\textwidth,
  colback=white,
  colframe=PromptBlack,
  colbacktitle=PromptBlack,
  coltitle=white,
  title=\bfseries #1,
  fonttitle=\normalsize,
  arc=2pt,
  outer arc=2pt,
  boxrule=0.9pt,
  top=4pt,
  bottom=4pt,
  left=0pt,
  right=6pt,
  listing options={
    basicstyle=\ttfamily\scriptsize,
    breaklines=true,
    breakatwhitespace=false,
    breakautoindent=false,
    breakindent=0pt,
    prebreak={},
    postbreak={},
    columns=fullflexible,
    keepspaces=true,
    showstringspaces=false,
    upquote=true,
    numbers=none
  }
}

\title{WebSwarm: Recursive Multi-Agent Orchestration for Deep-and-Wide Web Search}
\author{
    Xiaoshuai Song\textsuperscript{\rm 1}\thanks{Work done during internship at Kuaishou, supervised by Kangzhi Zhao (kangzhi.zhao@outlook.com).},
    Liancheng Zhang\textsuperscript{\rm 1}\textsuperscript{$*$},
    Kangzhi Zhao\textsuperscript{\rm 2}\corresponding,
    Yutao Zhu\textsuperscript{\rm 1},
    Zhongyuan Wang\textsuperscript{\rm 2}\textsuperscript{$*$},
    Guanting Dong\textsuperscript{\rm 1},
    Jinghan Yang\textsuperscript{\rm 2}\textsuperscript{$*$},
    Han Li\textsuperscript{\rm 2},
    Kun Gai\textsuperscript{\rm 2},
    Ji-Rong Wen\textsuperscript{\rm 1},
    Zhicheng Dou\textsuperscript{\rm 1}\corresponding
}
\affiliations{
    \textsuperscript{\rm 1}Gaoling School of Artificial Intelligence, Renmin University of China
    \textsuperscript{\rm 2}Kuaishou Technology\\
    {\{songxiaoshuai, dou\}@ruc.edu.cn, kangzhi.zhao@outlook.com} \\

}

\begin{document}

\maketitle

\begin{abstract}
Large language model (LLM)-based web search agents are transforming information seeking from simple factoid question answering into complex, deep-and-wide search and research-oriented tasks.
A single ReAct-style agent is constrained by one long trajectory and limited context, making it difficult to handle depth and coverage simultaneously. Existing multi-agent systems improve search coverage through parallel execution and aggregation, but still exhibit clear limitations in recursive depth, collaboration adaptability, and evidence-grounded expansion.
We propose WebSwarm, a progressive recursive delegation framework that jointly constructs task decomposition, recursive expansion, and agent collaboration during inference. 
WebSwarm dynamically instantiates agentic search nodes, each coupling a local objective with a search mode that specifies how the node should organize search and collaboration.
Each node can either solve its objective itself or further delegate child nodes; after solving, it returns evidence and results upward, enabling parent nodes to further expand, revise, or aggregate the search process.
To guide this process, WebSwarm first probes how task-relevant information is organized on the web to ground subsequent node expansion, and reuses process-level experience across homogeneous sibling nodes.
Experiments on BrowseComp-Plus, WideSearch, DeepWideSearch, and GISA show that WebSwarm consistently outperforms single-agent and multi-agent baselines on deep, wide, and interleaved deep-and-wide tasks. Further analyses of ablation, task difficulty, web tool efficiency, and model generalization explain WebSwarm's effectiveness and provide insights for multi-agent search systems\footnote{Github: \url{https://github.com/songxiaoshuai/WebSwarm}}.
\end{abstract}

\section{Introduction}
\label{sec:intro}

\begin{figure}[!t]
\centering
\includegraphics[width=1.0\columnwidth]{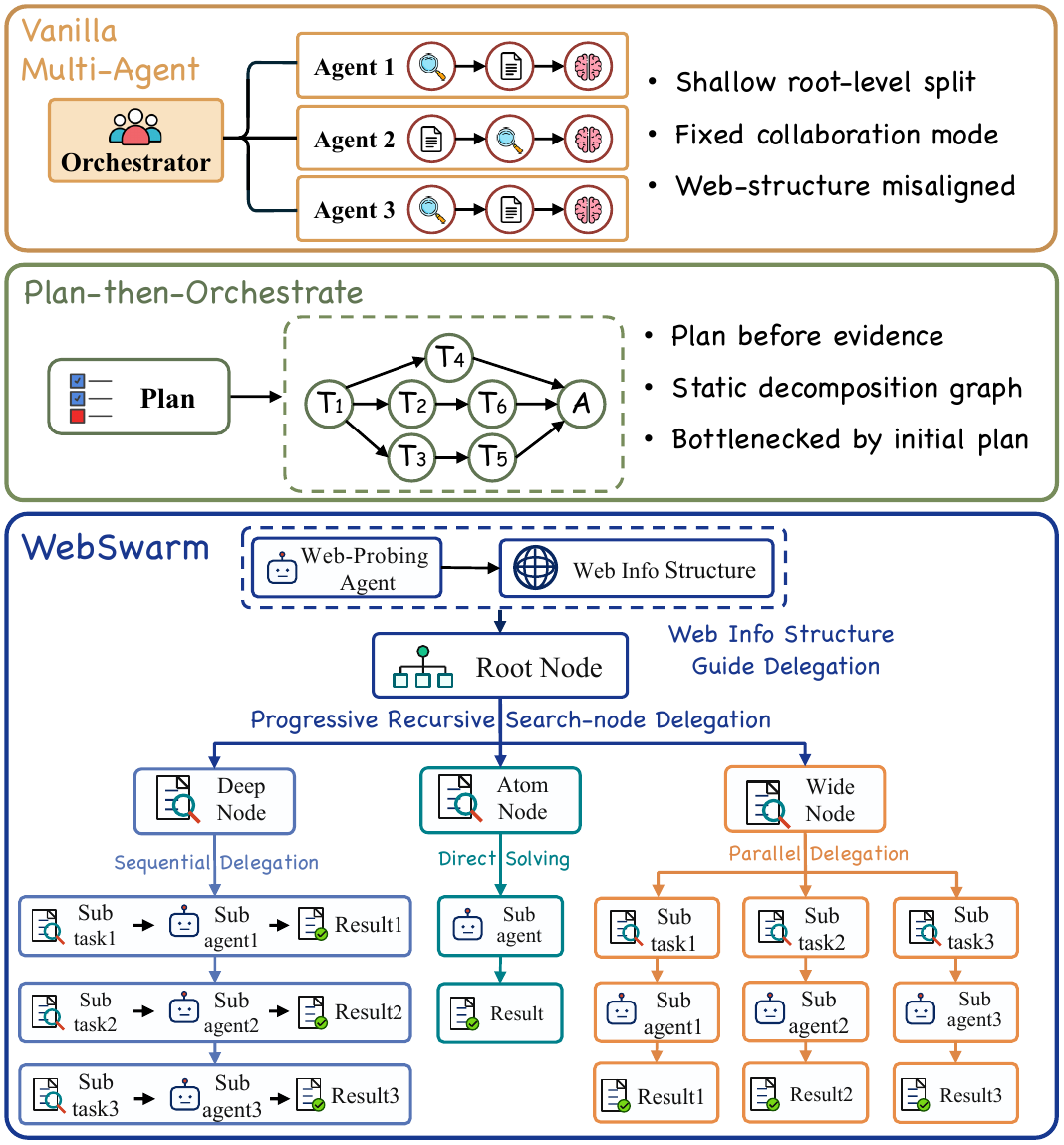}
\caption{Illustration of representative multi-agent orchestration paradigms and WebSwarm.}
\label{fig:method_intro}
\end{figure}

The development of large language models (LLMs) is driving web information seeking toward agentic search, enabling search agents to autonomously perform multi-turn search and web browsing to gather information for user queries~\citep{LLM-IR-Survey,SearchAgentSurvey}. As this paradigm evolves, search agents are moving beyond simple factoid QA toward more complex information-seeking tasks, such as research-level search and report-oriented investigation. This requires agents to support both deep and wide search: deep search resolves multi-hop dependencies and constraints, while wide search maintains sufficient coverage across candidate entities, web pages, and information sources. Recently, a series of benchmarks have evaluated the capability boundaries of search agents from the perspectives of depth, width, and their nested interaction~\citep{BrowseComp,BrowseComp-Plus,WideSearch,DeepWideSearch,GISA}.

Both benchmark results and practical experience show that a single ReAct~\citep{ReAct} agent struggles as task depth and width increase.  
To address this issue, recent studies have introduced multi-agent search systems~\citep{HiRA,Kimi-K2.5,Table-as-Search,A-MapReduce,ROMA,InfoSeeker,SearchSwarm}. These systems assign work to multiple agents and use mechanisms such as parallel search, cross-checking, and result aggregation to improve coverage and reliability.

For complex web search tasks with intertwined depth and width, the solving process is often difficult to fully determine from the initial query alone; instead, intermediate search evidence progressively reveals new entities, constraints, and goals. Therefore, rather than committing upfront to a fixed subtask decomposition and agent-collaboration structure, \textbf{an ideal multi-agent system should progressively and recursively construct both as evidence accumulates}. However, as illustrated in Figure~\ref{fig:method_intro}, existing multi-agent search systems still exhibit clear limitations in recursive depth, collaboration adaptability, and evidence-grounded expansion:

\textbf{(1) Shallow recursive depth.} 
Deeper subtasks may only become clear after earlier subtasks are expanded or solved, and some subtasks may introduce new dependencies and requirements. Thus, the task tree should support progressive and recursive expansion. Existing systems usually decompose only at the root level. When a task requires multi-level expansion, such as ``year$\rightarrow$brand$\rightarrow$model$\rightarrow$attribute'', deeper structures are forced into a long ReAct trajectory of a sub-agent, making the process close to single-agent search.

\textbf{(2) Limited collaboration adaptability.}
As the search process recursively unfolds, local search nodes may expose different goals and bottlenecks: fact lookup requires fast and reliable execution, wide search requires coverage over many items, deep search requires iterative clue discovery and verification, and open-set enumeration requires balancing recall and precision under unknown set boundaries. 
However, existing methods typically rely on a single global collaboration paradigm, such as multi-sample aggregation, serial handoff, or parallel divide-and-conquer. Although each paradigm is effective in certain search scenarios, a fixed collaboration pattern struggles to cover all  search needs or flexibly switch collaboration forms across different subtasks.

\textbf{(3) Weakly evidence-grounded expansion.}
Required information may be concentrated in a few aggregate pages, or distributed across timelines, entities, event sets, or attribute dimensions. A proper decomposition should align with the actual organization of web information. Therefore, the system must decide how to expand with respect to how relevant information is organized on the web. However, existing multi-agent systems often split tasks only by surface semantics  of the query. When the decomposition dimension is not aligned with the web information structure, the system may over-decompose concentrated information or split dispersed information along the wrong dimension, leading to redundant retrieval, insufficient recall, and difficult aggregation.

To address these limitations, we propose WebSwarm, a progressive recursive multi-agent framework that organizes complex web information seeking by dynamically creating and delegating search nodes with diverse local objectives and collaboration patterns.
In WebSwarm, the root agent receives the original task, creates and delegates search nodes. Each search node is itself an agent, receiving a local objective and a search mode. The search mode determines how the node solves its local objective: either by conducting iterative search on its own, or by recursively generating and delegating child nodes while organizing corresponding multi-node collaboration structures, such as parallel divide-and-conquer, sequential search-and-verification, or multi-path sampling and aggregation. Once a node completes its local objective, it returns the result upward, and the upper-level agent decides whether to further expand, revise, or terminate the search process based on the returned information. 
In this way, WebSwarm unifies recursive delegation, multi-level feedback, and diverse local collaboration structures into a progressive search process.
To avoid blind recursive delegation, WebSwarm further introduces two complementary signals: external web information structure and internal experience. On the one hand, the system performs lightweight probing of web information structure to determine whether relevant information is concentrated in a few aggregated pages or dispersed along a certain organizational dimension, thereby guiding how subsequent search nodes should be expanded. On the other hand, for homogeneous search nodes under the same parent node, WebSwarm distills trajectory experience from a small number of preceding nodes to guide the local search of subsequent nodes.

We evaluate WebSwarm on four challenging web information seeking benchmarks: BrowseComp-Plus, WideSearch, DeepWideSearch, and GISA, covering deep, wide, and hybrid deep-wide search tasks. Experimental results show that WebSwarm consistently outperforms ReAct agent and multi-agent baselines. Further analyses examine module ablations, the relationship between task difficulty and method performance, web tool-use efficiency, and generalization across different models, providing insights into WebSwarm's effectiveness and the design of multi-agent search systems.

Overall, we propose WebSwarm for deep and wide web search tasks. Our contributions are threefold:
\begin{itemize}
\item We propose a progressive recursive delegation framework that instantiates agentic search nodes coupling local objectives with search modes, jointly constructing decomposition, expansion, and collaboration.
\item We introduce web-structure-guided recursive delegation, grounding expansion in web information organization and further transferring subtask experience across homogeneous search nodes.
\item Experiments on four benchmarks verify WebSwarm’s effectiveness on complex web information seeking tasks involving deep, wide, and depth-width interleaved search.
\end{itemize}

\begin{figure*}[t]
\centering
\includegraphics[width=1.0\textwidth]{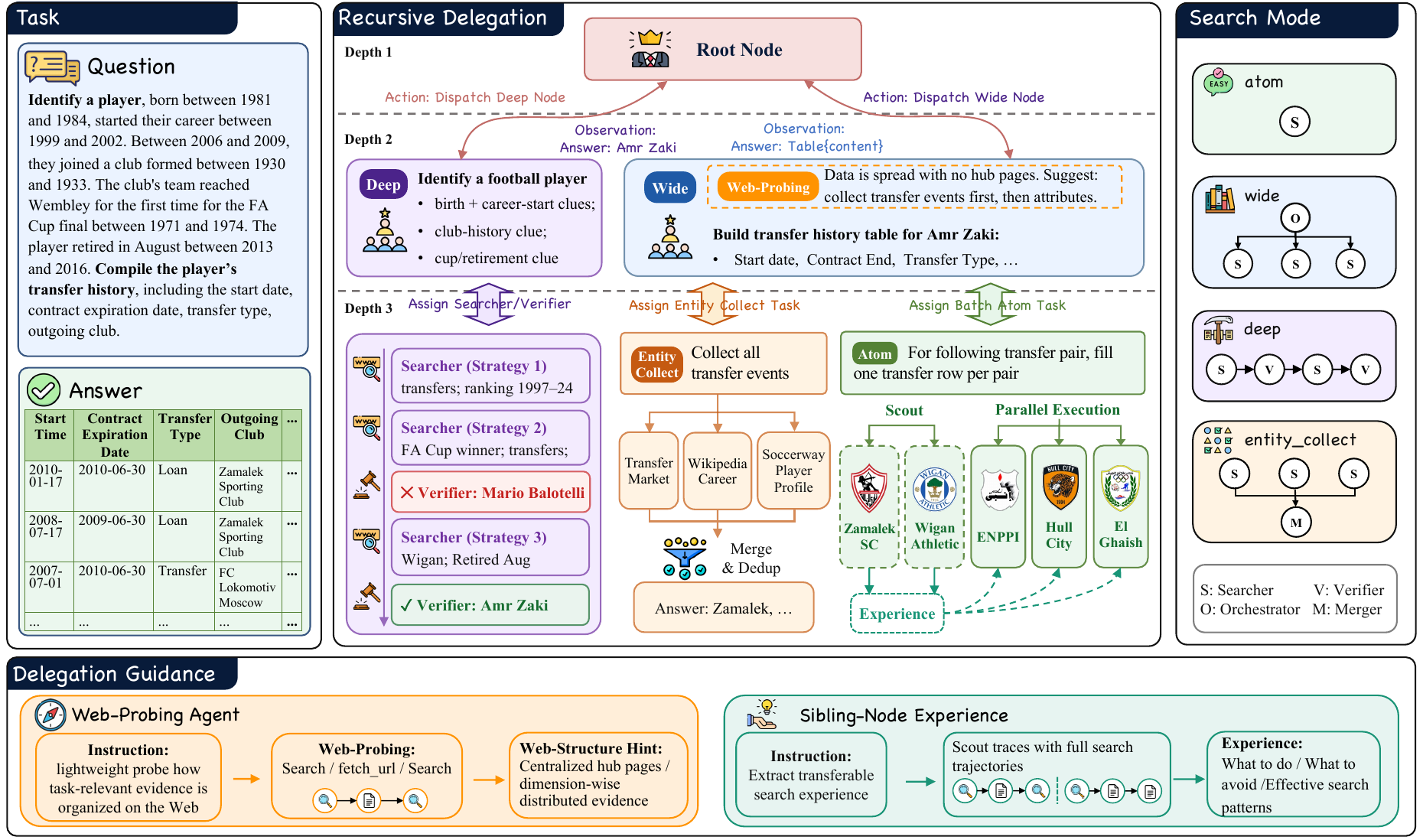} 
\caption{Overview of WebSwarm with a running example from the DeepWideSearch benchmark.}
\label{fig:method_overall}
\end{figure*}

\section{Related Work}
\label{sec:related_work}

\subsection{Agent for Web Information Seeking}
Agentic search combines LLMs with web search and browsing tools to solve information-seeking tasks~\citep{WebGPT,LLM-IR-Survey,Search-o1,SearchAgentSurvey}.
One line of work focuses on model training to internalize information-seeking strategies~\citep{Search-r1,WebThinker,Webdancer,TongyiDeepResearch}.
Recent benchmarks including BrowseComp~\citep{BrowseComp}, WideSearch~\citep{WideSearch}, DeepWideSearch~\citep{DeepWideSearch}, and GISA~\citep{GISA} reveal the difficulties that agents face in deep and wide web search tasks.
To address these challenges, inference-time architectures organize complex search through explicit state structures and multi-agent collaboration~\citep{Flash-Searcher,WideSeek-R1,SearchSwarm}.
Table-as-Search~\citep{Table-as-Search} and Web2BigTable~\citep{Web2BigTable} formulate search as table completion.
MindSearch~\citep{MindSearch} and HiRA~\citep{HiRA} adopt graph-based or hierarchical planning.
InfoSeeker~\citep{InfoSeeker} and A-MapReduce~\citep{A-MapReduce} emphasize hierarchical parallelism or MapReduce-style horizontal decomposition.
In contrast, WebSwarm formulates web search as evidence-driven recursive delegation: it instantiates search nodes as intermediate evidence reveals new local objectives, grounds their expansion in web information structure, and couples each objective with a search mode to handle heterogeneous local search needs.

\subsection{Multi-Agent Systems for Long-Horizon Tasks}
Early multi-agent systems, including CAMEL~\citep{CAMEL}, AutoGen~\citep{AutoGen}, MetaGPT~\citep{MetaGPT}, and ChatDev~\citep{ChatDev}, explore agent collaboration from perspectives such as role-playing, software development, debate, and voting.
Recent work further investigates agent orchestration for long-horizon tasks. Claude Agent Team builds agent teams that can communicate and collaborate with each other~\citep{ClaudeAgentTeam}; and Kimi-Swarm decomposes complex tasks into multiple subproblems and executes multiple worker agents in parallel~\citep{Kimi-K2.5}. In addition, Magentic-One employs a central orchestrator to coordinate specialized agents~\citep{Magentic-One}; ROMA~\citep{ROMA} represents long-horizon tasks as recursive subtask trees; AgentFugue~\citep{AgentFugue} reuses intermediate findings from parallel agents through a shared reasoning hub; and AggAgent~\citep{AggAgent} enables test-time scaling by aggregating multiple long-horizon agent trajectories.
In contrast, WebSwarm views multi-agent orchestration as recursive delegation rather than a fixed global collaboration topology. It recursively instantiates agentic nodes whose modes define local collaboration protocols; each node may solve locally, delegate child nodes, and return evidence upward, allowing decomposition and collaboration to co-evolve during inference.

\section{Methodology}
\label{sec:method}

\subsection{Preliminaries and Overview}
\label{subsec:method_overview}

\paragraph{ReAct Searcher.}
For a web information-seeking task $q_0$, the ReAct agent searches and reads web pages through web tools at each step until it returns the final answer $a$. The tool actions include \texttt{search(query)} and \texttt{fetch\_url(url)}: the former returns the most relevant webpage URLs and snippets for the query, while the latter returns the text content of a specified webpage.
\paragraph{WebSwarm.}
As shown in Figure~\ref{fig:method_overall}, at the core of WebSwarm is a recursive delegation process, where the system progressively solves the original task by dynamically creating search nodes and enabling delegation and feedback among them. Each search node is itself an agent, receiving a local objective and a search mode, where the search mode determines whether the node directly solves the current objective or further delegates child nodes and organizes local collaboration structures. The results returned by child nodes then serve as the basis for upper-level nodes to further expand, revise, or aggregate the search process (§3.2).
To make recursive delegation more reliable and efficient, WebSwarm further leverages two types of guiding signals: probed external web information structure guides how subsequent search nodes are expanded, while internal process experience guides the subsequent search of homogeneous nodes (§3.3).

\subsection{Recursive Search Delegation}
Given the original task $q_0$, WebSwarm dynamically constructs a recursive delegation tree $\mathcal{T}=(\mathcal{V},\mathcal{E})$ during execution. The root node $v_0$ receives $q_0$ and serves as the starting point of recursive delegation. Each non-root node $v\in\mathcal{V}$ corresponds to an agent and receives a local objective $q_v$ and a search mode $m_v$: $q_v$ specifies what the node should solve, while $m_v$ specifies how it organizes search and collaboration. An edge $(u,v)\in\mathcal{E}$ denotes one delegation from parent node $u$ to child node $v$. During solving, node $v$ maintains the set of results $R_v$ returned by its children, and returns result $r_v$ to its parent after completing its local objective.

\subsubsection{Objective Decomposition and Delegation.}
The basic action in recursive delegation is that a node creates one or more child nodes. For a node $v$, the agent determines whether to expand further based on the local objective $q_v$, search mode $m_v$, and collected child results $R_v$. If the current evidence is insufficient, node $v$ generates a set of child delegations:
\begin{equation}
    C_v=\{(q_i,m_i)\}_{i=1}^{n_v}.
\end{equation}
Each child delegation specifies both a local objective $q_i$ and a corresponding search mode $m_i$, instantiating a new search node $v_i$ and forming an edge $(v,v_i)$ in the recursive delegation tree.
The key point is that WebSwarm generates not merely a set of subtasks, but a set of objective–mode pairs. Since different local objectives often require different solution strategies, coupling the objective with its search mode during delegation ensures each downstream node has a solving protocol and collaboration structure suited to its goal. 

\subsubsection{Search Node Solving.}
The search mode specifies the solving protocol for the subtask $q_v$. By assigning different instructions, roles, and tools to a node, the search mode determines whether the node directly calls web tools, or further creates child nodes and organizes collaboration among them. 
WebSwarm defines four search modes:
\begin{equation}
\mathcal{M}=\{\textit{atom},\textit{deep},\textit{wide},\textit{entity\_collect}\}.
\end{equation}
They correspond to four subtask bottlenecks in web search:
\begin{itemize}
% atom
\item \textbf{\textit{atom}: atomic fact lookup.}
\textit{atom} is the basic search mode, suited to focused, narrow local queries. Its goal is to quickly and reliably find evidence. Such a node corresponds to a ReAct agent that directly calls web tools to locate relevant evidence and return the result.
% deep
\item \textbf{\textit{deep}: iterative search and verification.}
\textit{deep} applies to unknown-target identification or multi-constraint reasoning. The challenge is not covering many objects, but iteratively proposing candidates from indirect clues, verifying them, and adjusting the search direction. Thus, \textit{deep} uses a serial search–verification structure: iteratively dispatches searchers and verifiers. Searchers propose candidates or evidence from different clue paths, while verifiers independently check whether candidates satisfy the constraints. Each searcher or verifier is an \textit{atom} node.
% wide
\item \textbf{\textit{wide}: parallel divide-and-conquer.} 
\textit{wide} is suited to collecting similar information over a set of objects or dimensions, such as filling attributes for multiple entities. The bottleneck is coverage and structured aggregation, so \textit{wide} concurrently dispatches a set of independent child nodes and merges their results. Its children are not limited to \textit{atom}: in tasks combining depth and breadth, child nodes can also be \textit{deep}, \textit{entity\_collect}, or  \textit{wide}, enabling nested information collection.
% entity collect
\item \textbf{\textit{entity\_collect}: multi-path recall and verification.}
\textit{entity\_collect} applies when the set boundary is unknown and the task is to enumerate a complete set of entities. Unlike attribute filling over known objects, its core challenge is balancing recall and precision: a single search path may miss members, while broadening the search directly may introduce noise. 
Therefore, \textit{entity\_collect} concurrently dispatches multiple \textit{atom} search nodes to recall candidate entities from different sources or perspectives, and obtains the target set by merging, deduplicating, and verifying low-confidence candidates.
\end{itemize}
Through these four search modes, WebSwarm enables each search node to choose an appropriate solving protocol for its local objective and combine multiple collaboration structures within the same recursive process.

\subsubsection{Evidence Update and Continuation.}
When a child node $v_i$ completes its local objective, it returns the result $r_i$ to its parent node $v$, which updates its evidence state:
\begin{equation}
R_v \leftarrow R_v \cup \{r_i\}.
\end{equation}
Based on these returned results $R_v$, the parent decides whether to continue or stop. If the local objective remains unresolved, it may generate new sub-delegations, revise the search direction, or request finer-grained evidence. Otherwise, it aggregates the collected results and returns upward:
\begin{equation}
r_v=\textsc{Aggregate}(q_v,R_v).
\end{equation}
Thus, returned results serve not only as answer components but also as control signals for further delegation. WebSwarm therefore alternates between top-down delegation and bottom-up evidence feedback, allowing the search structure to unfold progressively during execution.

\subsection{Guiding Delegation with Web Structure and Experience}
During recursive delegation, agent nodes may generate misaligned expansions, redundant delegations, or repeated trial-and-error during search. To improve reliability and efficiency, WebSwarm introduces two complementary guidance signals: web-structure cues $h_v$ from lightweight probing, which constrain the scope, granularity, and dimensions of subsequent node expansion, and internal search experience $k_v$, which guides the solving process of homogeneous nodes.

\subsubsection{Web-Structure-Guided Expansion.}
Recursive expansion requires deciding not only whether to generate more search nodes, but also along which dimension to expand. Prior methods mainly rely on LLM agents to decompose tasks based on query surface semantics, which can be misaligned with the actual organization of web evidence in two ways: (1) when the required information is concentrated in a few aggregate pages, premature expansion by entity or attribute causes redundant retrieval; (2) when information is dispersed across sources organized by external structures such as time, events, organizations, or entities, choosing the wrong expansion dimension reduces efficiency and increases aggregation noise.

To mitigate this misalignment, WebSwarm introduces a Web-Probing Agent for nodes that require wide expansion. It acts as a lightweight pre-expansion structure scout, probes how task-relevant evidence is organized on the web and uses this structure signal to guide subsequent delegation.
Given the current local objective $q_v$, the Web-Probing Agent performs lightweight search and webpage reading, and returns a web-structure hint:
\begin{equation}
    h_v = \textsc{WebProbing}(q_v).
\end{equation}
Here, $h_v$ summarizes the preliminarily observed evidence distribution, representative pages and their supporting signals, and possible expansion axes. The node then generates child delegations conditioned on $h_v$:
\begin{equation}
    C_v=\textsc{Delegate}(q_v,m_v,R_v,h_v).
\end{equation}
When information is concentrated in a few aggregate pages, WebSwarm creates a small number of extraction-oriented search nodes around these pages to avoid redundant parallelism. When evidence is dispersed across multiple sources, it instead expands along the web-organization axes suggested by $h_v$, improving coverage and aggregability. 

% Subtask Experience.
\subsubsection{Experience-Guided Node Solving.}
During recursive delegation, the \textit{wide} node often creates a batch of homogeneous search nodes, such as collecting attributes for a set of known entities. Although these nodes target different entities, they often share similar query patterns, reliable sources, page structures, and failure paths. If each node searches in isolation, it may repeat unreliable strategies, miss useful source patterns, or produce inconsistent results across sibling nodes. 
To address this, WebSwarm introduces within-instance subtask experience transfer: it first executes a small number of scout nodes, extracts process-level subtask experience from their search trajectories, and uses it to guide subsequent homogeneous nodes toward more reliable solving.
Specifically, given a set of homogeneous child delegations $C_v$ under node $v$, WebSwarm first executes a small scout subset $C_v^{s}\subset C_v$:
\begin{equation}
    (r_i,\tau_i)=\textsc{SolveNode}(v_i), \quad (q_i,m_i)\in C_v^{s}.
\end{equation}
where $v_i$ is the search node instantiated by the child delegation $(q_i,m_i)$, and $\tau_i$ denotes its solving trajectory. WebSwarm then extracts subtask experience $k_v$ from these trajectories, including useful query patterns, reliable sources, page types, and invalid paths:
\begin{equation}
    k_v=\textsc{ExtractExperience}(\{\tau_i:(q_i,m_i)\in C_v^{s}\}).
\end{equation}
For the remaining sibling nodes, WebSwarm injects $k_v$ into their solving context:
\begin{equation}
    r_j=\textsc{SolveNode}(v_j;k_v), \quad (q_j,m_j)\in C_v\setminus C_v^{s}.
\end{equation}
% Here, $k_v$ provides process-level guidance rather than answer-level facts specific to any scout node. 
Different from studies focusing on cross-task experience or skills~\citep{A-MapReduce,EvoSkill}, this mechanism only reuses experience within the same instance and among homogeneous sibling nodes under the same parent, thereby preserving evaluation-sample independence.

\begin{table*}[t]
\centering
\small
\setlength{\tabcolsep}{1.1mm}        % 列间距
\renewcommand{\arraystretch}{1.2}   % 行间距
\begin{tabular}{llcccccccccccc}
\toprule
\textbf{Method}
& \textbf{Backbone}
& \textbf{BC-Plus}
& \multicolumn{3}{c}{\textbf{WideSearch-EN}} 
& \multicolumn{3}{c}{\textbf{DeepWideSearch-EN}} 
& \multicolumn{5}{c}{\textbf{GISA}} \\
\cmidrule(lr){3-3} \cmidrule(lr){4-6} \cmidrule(lr){7-9} \cmidrule(lr){10-14}
~
& ~
& ACC & SR & Row F1 & Item F1
& SR & Row F1 & Item F1
& Item & Set & List & Table & Overall \\
\midrule
\multicolumn{14}{l}{\textbf{\textit{Single-Agent}}} \\
ReAct & Kimi-K2 
& \underline{66.50}
& \textbf{7.00} & 37.42 & 67.65
& 3.95 & 27.58 & 53.57
& 18.18 & 57.18 & 48.76 & 58.34 & 54.58 \\
ReAct & Qwen3.5-35B
& 56.50 & 5.00 & 37.38 & 64.82
& 3.95 & \underline{28.79} & 52.46 
& 27.27 & 55.76 & 51.72 & 56.03 & 53.74 \\
ReAct & Qwen3-235B 
& 21.50
& 3.00 & 20.89 & 49.15
& 0.00 & 11.75 & 36.12
& \textbf{40.91} & 52.37 & 36.48 & 43.93 & 45.57 \\
ReAct & GLM-4.5
& 50.50 & 4.00 & 33.23 & 64.61
& 3.95 & 20.08 & 46.63
& 27.27 & 51.27 & 50.58 & 59.78 & 55.54 \\
\midrule
\multicolumn{14}{l}{\textbf{\textit{Multi-Agent}}} \\
Swarm-Agent & GLM-4.5
& 64.50
& \underline{6.00} & 36.66 & 68.79
& 1.32 & 27.03 & 51.41
& \underline{31.82} & 54.43 & 52.33 & 60.66 & 57.05 \\
Flash-Searcher & GLM-4.5
& 54.00 & 5.00 & 36.53 & 68.68
& \underline{5.26} & 25.17 & 50.00
& 18.18 & 54.91 & 56.85 & 56.72 & 54.22 \\
Table-as-Search & GLM-4.5
& 62.50 & \underline{6.00} & 37.97 & 69.44
& 3.95 & 23.12 & 54.96
& 22.73 & \underline{60.44} & \underline{57.86} & \underline{60.82} & 58.14 \\
ROMA & GLM-4.5
& 42.50 
& 5.00 & 33.42 & 67.19
& 2.63 & 24.02 & 50.56
& 27.27 & 56.91 & 56.46 & 60.55 & 57.57 \\
InfoSeeker & GLM-4.5
& 59.50 & 4.00 & \underline{39.98} & \underline{71.91}
& 2.63 & 25.81 & \underline{55.10}
& 27.27 & 54.58 & 57.31 & 40.82 & \underline{58.99} \\
\rowcolor{rowblue}[2\tabcolsep][2\tabcolsep]
\textbf{WebSwarm} & GLM-4.5
& \makecell[c]{\textbf{68.00}\\{\scriptsize (+17.50)}}
& \makecell[c]{\textbf{7.00}\\{\scriptsize (+3.00)}}
& \makecell[c]{\textbf{44.14}\\{\scriptsize (+10.91)}}
& \makecell[c]{\textbf{74.37}\\{\scriptsize (+9.76)}}
& \makecell[c]{\textbf{6.58}\\{\scriptsize (+2.63)}}
& \makecell[c]{\textbf{29.64}\\{\scriptsize (+9.56)}}
& \makecell[c]{\textbf{58.40}\\{\scriptsize (+11.77)}}
& \makecell[c]{\textbf{40.91}\\{\scriptsize (+13.64)}}
& \makecell[c]{\textbf{61.03}\\{\scriptsize (+9.76)}}
& \makecell[c]{\textbf{66.04}\\{\scriptsize (+15.46)}}
& \makecell[c]{\textbf{63.69}\\{\scriptsize (+3.91)}}
& \makecell[c]{\textbf{62.30}\\{\scriptsize (+6.76)}} \\
\bottomrule
\end{tabular}
\caption{Performance comparison of methods across four benchmarks. The top two results are highlighted in bold and underlined. Numbers in parentheses denote the gain of WebSwarm compared to the ReAct agent under GLM-4.5. BC-Plus: BrowseComp-Plus; SR: Success Rate; Qwen3-235B: Qwen3-235B-A22B-2507; Kimi-K2: Kimi-K2-Thinking.}
\label{tab:main_results}
\end{table*}

\section{Experiment}
\label{sec:experiment}

\subsection{Experimental Setup}
\label{subsec:experiment_setting}

\subsubsection{Benchmarks.}
We evaluate WebSwarm on four challenging web information-seeking benchmarks: BrowseComp-Plus~\citep{BrowseComp-Plus} for deep factual search, WideSearch~\citep{WideSearch} for structured wide information collection, DeepWideSearch~\citep{DeepWideSearch} for nested deep and wide search, and GISA~\citep{GISA} for general information seeking. For resource efficiency, we randomly sample 200 instances from BrowseComp-Plus and use the English subsets of WideSearch and DeepWideSearch.

Following the official evaluation settings, we report answer accuracy (ACC) for BrowseComp-Plus. For WideSearch and DeepWideSearch, whose outputs are tables, we report item-F1, row-F1, and success rate (SR) to measure cell-level, row-level, and full-table correctness, respectively. GISA covers diverse task types and metrics, including exact match (EM) for items, set F1, list F1, and table item-F1.

\subsubsection{Baselines.}
We compare WebSwarm with the following representative search agent and multi-agent methods: ReAct agent~\citep{ReAct}; Swarm-Agent, which follows the Kimi-Swarm~\cite{Kimi-K2.5} and dynamically creates and assigns tasks to subagents; Flash-Searcher~\citep{Flash-Searcher}, which organizes subtasks as a dynamic dependency graph; Table-as-Search~\citep{Table-as-Search}, which represents search progress as table completion; InfoSeeker~\citep{InfoSeeker}, which adopts a hierarchical Host--Manager--Worker architecture; and ROMA~\citep{ROMA}, which recursively decomposes tasks and aggregates leaf-level results.

\subsubsection{Implementation.}
We mainly use GLM-4.5 as the backbone model for WebSwarm and multi-agent baselines, and compare different models in Section 4.5. For web interaction, all methods use the same Web-Search and Page-Browse tools. 
For WebSwarm, we set the number of parallel search paths in \textit{entity\_collect} to 3, and use 2 scout child nodes for experience extraction.
We use each model's maximum context length and allow up to 200 action steps to avoid prematurely truncating agents. 
We provide detailed descriptions of the benchmarks, baselines, implementation in Appendix.

\subsection{Main Results}
\label{subsec:main_results}

As shown in Table~\ref{tab:main_results}, WebSwarm achieves the best or competitive results across four benchmarks, consistently outperforming both single-agent ReAct and multi-agent baselines. Next, we analyze the results from three aspects:

\textbf{(1) Deep Search.}
On BrowseComp-Plus, WebSwarm improves over ReAct by 17.50 accuracy points and outperforms the strongest multi-agent baseline by 3.50 points. 
This gain is primarily attributed to the \textit{deep} search mode, where the searcher agent iteratively shifts exploration perspectives and refines candidate hypotheses in a sequential manner, while an independent verifier adversarially evaluates and filters candidate answers, leading to deeper web information exploration.

\textbf{(2) Wide Search.}
On WideSearch-EN, WebSwarm achieves gains of 10.91 and 9.76 points in Row F1 and Item F1 over ReAct, and also shows improvements over multi-agent baselines.
These results show that WebSwarm is effective not only at expanding search coverage, but also at aggregating structured evidence with higher row- and item-level correctness. 
These gains stem from recursive delegation among \textit{entity\_collect}, \textit{wide}, and \textit{atom} search nodes, further supported by Web-structure-guided expansion and process-level experience reuse across homogeneous sibling nodes.

\textbf{(3) Interleaved Deep and Wide Search.}
WebSwarm improves over ReAct by 9.56 Row F1 and 11.77 Item F1 on DeepWideSearch-EN. On GISA, WebSwarm performs strongly across all task subsets.
These results indicate that when tasks require alternating between deep and wide exploration, or combining both within a single workflow, fixed collaboration patterns may be less effective at coordinating different search phases. WebSwarm addresses this through a recursive delegation structure and mode-guided node solving, allowing flexible transitions between deep and wide search as evidence accumulates.

\begin{table}[t]
\centering
\small
\setlength{\tabcolsep}{1.2mm}        % 列间距
\renewcommand{\arraystretch}{1.2}   % 行间距
\begin{tabular}{lccc}
\toprule
\textbf{Method}
& \multicolumn{1}{c}{\textbf{BC-Plus}}
& \multicolumn{1}{c}{\textbf{WideSearch}}
& \multicolumn{1}{c}{\textbf{DeepWideSearch}} \\
\cmidrule(lr){2-2} \cmidrule(lr){3-3} \cmidrule(lr){4-4}
~
& \textbf{ACC}
& \textbf{Item F1}
& \textbf{Item F1} \\
\midrule
\textbf{WebSwarm} 
& \textbf{68.00} & \textbf{74.37} & \textbf{58.40} \\
\quad w/o recursive 
& 63.50 & 68.38 & 55.79  \\
\quad All-to-\textit{wide}
& 63.00 & 72.01 & 55.87 \\
\quad All-to-\textit{deep} 
& 67.50 & 69.94 & 54.51 \\
\bottomrule
\end{tabular}
\caption{Ablation of recursive delegation and search mode.}
\label{tab:structure_ablation}
\end{table}

\begin{table}[t]
\centering
\small
\setlength{\tabcolsep}{1.3mm}        % 列间距
\renewcommand{\arraystretch}{1.2}   % 行间距
\begin{tabular}{lcccc}
\toprule
\textbf{Method}
& \multicolumn{2}{c}{\textbf{WideSearch}}
& \multicolumn{2}{c}{\textbf{DeepWideSearch}} \\
\cmidrule(lr){2-3} \cmidrule(lr){4-5}
~
& Item F1 & Web Tool
& Item F1 & Web Tool \\
\midrule
\textbf{WebSwarm} 
& 74.37	& 137.03 &	58.40	& 203.73 \\
\quad w/o Web-Probing
& 74.90 & 239.90 & 58.93 & 331.39 \\
\quad w/o Experience
& 71.20 & 132.57 & 55.48 & 220.07  \\
\bottomrule
\end{tabular}
\caption{Ablation of Web-probing and experience reuse.}
\label{tab:experience_ablation}
\end{table}

\begin{figure*}[t]
\centering
\includegraphics[width=1.0\textwidth]{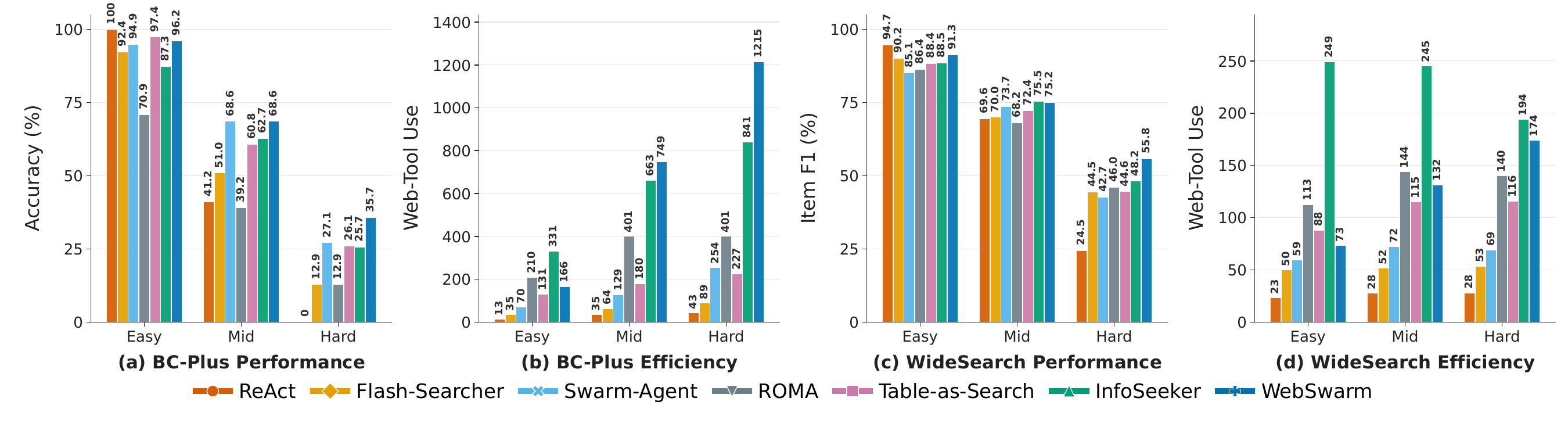}
\caption{Performance and web tool usage of GLM-4.5 across difficulty levels on BrowseComp-Plus and WideSearch-EN.}
\label{fig:difficulty_analysis}
\end{figure*}

\subsection{Ablation Analysis}
\label{subsec:ablation_analysis}

\textbf{Structure Ablation.}
To analyze the role of recursive delegation and search mode, we construct three variants: \textit{w/o Recursive Delegation}, which allows the root node to create child nodes but prevents child nodes from further delegation; and All-to-\textit{wide} / All-to-\textit{deep}, which force all non-\textit{atom} nodes to use a single search mode.
As shown in Table~\ref{tab:structure_ablation}, removing recursive delegation consistently degrades performance, confirming the value of progressively refined search structures.
The single-mode variants show task-dependent behavior: All-to-\textit{deep} remains close to WebSwarm on BC-Plus, but drops clearly on WideSearch and DeepWideSearch; conversely, All-to-\textit{wide} is less harmful on WideSearch but substantially weakens BC-Plus. It highlights the importance of matching local objectives with suitable search modes.

\textbf{Guidance Signal Ablation.}
Since web-structure probing and experience reuse mainly affect expansion and local solving under \textit{wide} nodes, we conduct ablations on WideSearch and DeepWideSearch in Table~\ref{tab:experience_ablation}. Removing Web-Probing causes only minor changes in Item F1, but substantially increases the average number of Web tool calls from 137.03 to 239.90 and from 203.73 to 331.39, respectively. This suggests that Web-Probing primarily improves search efficiency by reducing redundant or misaligned node expansion. In contrast, removing experience reuse consistently lowers Item F1 on both datasets, indicating that process-level experience improves the reliability of homogeneous sibling nodes. Overall, Web-Probing and experience reuse play complementary roles: the former reduces exploration cost, while the latter improves node-solving quality.

\subsection{Task Difficulty and Tool Usage Analysis}
\label{subsec:difficulty_effiency_analysis}
To investigate WebSwarm's performance gains across task difficulties and its web-tool efficiency, we use ReAct Agent as the anchor for difficulty stratification. For BrowseComp-Plus, difficulty is defined by ReAct Agent's pass rate over three samples: Easy means all samples pass, Hard means all fail, and Mid denotes the remaining cases. For WideSearch-EN, samples are sorted by ReAct Agent's Item F1 score; the top group is treated as Easy, the bottom group as Hard, and the intermediate group as Mid.
As shown in Figure~\ref{fig:difficulty_analysis}(a) and Figure~\ref{fig:difficulty_analysis}(c), WebSwarm's advantage becomes more pronounced as task difficulty increases, especially on hard samples. On the Hard subset, WebSwarm substantially improves over ReAct Agent: from 0.0 to 35.7 on BrowseComp-Plus, and from 24.5 to 55.8 on WideSearch-EN. It also clearly outperforms other multi-agent baselines.
Figures~\ref{fig:difficulty_analysis}(b) and \ref{fig:difficulty_analysis}(d) show the corresponding web-tool usage. Multi-agent methods generally make more tool calls than ReAct Agent. WebSwarm adaptively increases resource use with task difficulty: it remains relatively controlled on Easy samples, while allocating substantially more search and reading budget to Hard samples. These results indicate that WebSwarm matches other multi-agent baselines on simple tasks while breaking the performance ceiling of ReAct Agent and multi-agent baselines on complex and long-tail samples.

\begin{table}[t]
\centering
\small
\setlength{\tabcolsep}{1.2mm}        % 列间距
\renewcommand{\arraystretch}{1.2}   % 行间距
\begin{tabular}{lccccc}
\toprule
\textbf{Method}
& \textbf{BC-Plus}
& \multicolumn{2}{c}{\textbf{WideSearch}}
& \multicolumn{2}{c}{\textbf{DeepWideSearch}} \\
\cmidrule(lr){2-2} \cmidrule(lr){3-4} \cmidrule(lr){5-6}
~
& \textbf{ACC}
& \textbf{Row F1} & \textbf{Item F1}
& \textbf{Row F1} & \textbf{Item F1} \\
\midrule
\multicolumn{6}{l}{\textbf{Qwen3-32B}} \\
ReAct & 12.00 & 5.43 & 29.11 & 3.28 & 20.53 \\ 
WebSwarm & 19.50 & 9.01 & 34.47 & 7.06 & 25.54 \\ 
\midrule 
\multicolumn{6}{l}{\textbf{Qwen3.5-35B}} \\ 
ReAct & 56.50 & 37.38 & 64.82 & 28.79 & 52.46 \\ 
WebSwarm & 61.00 & 47.54 & 75.91 & 35.90 & 58.34 \\
\bottomrule
\end{tabular}
\caption{Performance comparison across different models.}
\label{tab:different_model}
\end{table}

\subsection{Performance across Different LLMs}
\label{subsec:backbone_analysis}

To verify WebSwarm's applicability across different backbones, we conduct experiments with Qwen3-32B and Qwen3.5-35B, as shown in Table~\ref{tab:different_model}.
The ReAct baseline scores show that these backbones differ substantially in their native agentic search ability: Qwen3-32B is relatively weak, while Qwen3.5-35B already exhibits strong long-horizon web search capability.
Across both settings, WebSwarm consistently improves over ReAct on deep, wide, and deep-wide search tasks.
These results indicate that WebSwarm's gains are not tied to a specific backbone capability level.
For weaker models, recursive delegation provides additional structure for long-horizon search; for stronger models, mode-guided node solving, web-structure-guided expansion, and process-level experience reuse further improve search depth and coverage.

\section{Conclusion}
\label{sec:conculsion}
In this paper, we propose WebSwarm for complex web information-seeking tasks that require  deep reasoning and broad information coverage. Unlike static upfront decomposition or fixed collaboration patterns, WebSwarm formulates search as progressive recursive delegation over agentic search nodes. Each node couples a local objective with a search mode, may solve locally or delegate child nodes, and returns evidence upward to support further expansion, revision, or aggregation. By introducing Web-structure-guided expansion and experience reuse across homogeneous sibling nodes, WebSwarm jointly constructs task decomposition, recursive expansion, and agent collaboration as evidence accumulates. Experiments across multiple benchmarks show that WebSwarm outperforms both single-agent and multi-agent baselines on deep, wide, and interleaved deep-and-wide tasks, with particularly clear gains on difficult examples.

\bibliography{aaai2027}

\appendix

\section{Limitations}
\label{subapp:limitation}

In this paper, we propose WebSwarm, a progressive recursive delegation framework for complex web information-seeking tasks. Although WebSwarm improves both deep and wide search, it still has several limitations.
(1) In terms of research scope, WebSwarm mainly focuses on inference-time multi-agent orchestration, without involving data construction or policy optimization for multi-agent training. Therefore, node delegation, search-mode assignment, Web-structure-guided expansion, and subtask experience reuse still rely on the understanding and reasoning abilities of the base LLM, which may not fully unlock the model's potential for recursive multi-agent collaboration.
(2) In terms of resource consumption, although WebSwarm adaptively instantiates search nodes and expands the delegation structure according to task difficulty, it still requires more LLM calls and web-tool requests than a single ReAct agent, leading to higher inference cost and latency.
(3) In terms of modality, WebSwarm mainly targets information-seeking tasks based on text-based web tools and has not yet fully covered multimodal web information, such as images, videos, and audio.
Looking to the future, WebSwarm can be extended in two directions. First, multi-agent training can enable models to learn better recursive delegation, search-mode assignment, and collaboration strategies from search trajectories. Second, the current text-based web-tool framework can be expanded to multimodal and GUI-based web scenarios.

\section{Details of WebSwarm}
\label{subapp:webswarm_details}

\subsection{Details of Prompts}
Figure~\ref{fig:root-agent-prompt} shows the system prompt of the root agent, which instructs it to iteratively generate subtasks and select their execution types. Figure~\ref{fig:explore-agent-prompt} presents the system prompt of the web structure probing agent, which guides the agent to pre-interact with the web and understand how task-relevant information is organized online. Figure~\ref{fig:skill-extraction-prompt} shows the system prompt for subtask-skill summarization, which instructs the LLM to summarize reusable experience from completed subtask trajectories for subsequent subtasks.

\subsection{Pseudo-code}
We summarize the pseudo-code of WebSwarm inference in Algorithm 1.

\begin{algorithm}[!t]
\caption{WebSwarm Inference}
\label{alg:webswarm}
\begin{algorithmic}[1]
\Require user task $q_0$
\Ensure final answer $a$
\Auxiliary
$q_v$: local objective;
$m_v$: search mode;
$R_v$: child results;
$h_v$: web-structure hint;
$C_v$: child delegations;
$k$: injected experience;
$\tau_v$: solving trajectory.

\State $v_0 \gets \textsc{CreateNode}(q_0)$
\State $(a,\tau_{v_0}) \gets \textsc{SolveNode}(v_0,\emptyset)$
\State \Return $a$

\Function{SolveNode}{$v,k$}
    \State $R_v,\tau_v \gets \emptyset,\emptyset$
    \While{\textbf{not} $\textsc{Ready}(q_v,R_v)$}
        \State $h_v \gets \textsc{WebProbing}(q_v)$ if $\textsc{NeedProbe}(v,R_v)$ else $\emptyset$
        \State $(y,\tau_y) \gets \textsc{SolveOrDelegate}(q_v,m_v,R_v,h_v,k)$
        \State $\tau_v \gets \tau_v \circ \tau_y$
        \If{$y$ is a result}
            \State \Return $(y,\tau_v)$
        \EndIf

        \State $(C_v,k^s,R^s,T^s) \gets \textsc{ScoutExperience}(v,y)$
        \State $(R^c,T^c) \gets \textsc{RunChildren}(v,C_v,k^s)$
        \State $R_v \gets R_v \cup R^s \cup R^c$
    \EndWhile
    \State \Return $(\textsc{Aggregate}(q_v,R_v),\tau_v)$
\EndFunction

\Function{ScoutExperience}{$v,C$}
    \If{\textbf{not} $\textsc{NeedExperience}(C)$}
        \State \Return $(C,\emptyset,\emptyset,\emptyset)$
    \EndIf
    \State $C^s \gets \textsc{SelectScouts}(C)$
    \State $(R^s,T^s) \gets \textsc{RunChildren}(v,C^s,\emptyset)$
    \State \Return $(C\setminus C^s,\textsc{ExtractExperience}(T^s),R^s,T^s)$
\EndFunction

\Function{RunChildren}{$v,C,k$}
    \State $R,T \gets \emptyset,\emptyset$
    \ForAll{$(q_i,m_i)\in C$}
        \State $u_i \gets \textsc{CreateChild}(v,q_i,m_i)$
        \State $(r_i,\tau_i) \gets \textsc{SolveNode}(u_i,k)$
        \State $R \gets R\cup\{r_i\},\quad T \gets T\circ\tau_i$
    \EndFor
    \State \Return $(R,T)$
\EndFunction
\end{algorithmic}
\end{algorithm}

\section{Details of Experiments}
\label{app:exp_details}

\subsection{Details of Web Tools}
\label{subapp:web_tool_details}
WebSwarm and all baselines interact with the web through a unified tool interface:
\begin{itemize}
\item \textbf{Web-Search Tool}:
Given a search query, the Web-Search Tool calls the Serper API~\footnote{ https://serper.dev/} to retrieve candidate web pages. The returned results are converted into a structured format containing the title, URL, snippet, date, and other available metadata. To control the search space, the tool returns the top-5 search results for each query.
\item \textbf{Page-Browse Tool}:
Given a URL, the Page-Browse Tool uses the Jina Reader API~\footnote{https://jina.ai/reader} to fetch the page content in Markdown format, with the page length limited to 80K tokens. Since raw web pages can be long and noisy, we further use the same LLM as the experimental agent to generate a goal-conditioned summary of the page. The summarized content is then returned to the agent for evidence verification, table filling, or subsequent reasoning.
\end{itemize}
Figure~\ref{fig:web-tool-interface} shows the tool descriptions provided to agents.

\subsection{Details of Agent Inference}
For all locally deployed models running on NVIDIA A800 GPUs, we use SGLang~\footnote{https://github.com/sgl-project/sglang} as the inference engine and keep the same hyperparameters throughout the experiments: thinking mode is enabled, temperature is set to $0.6$, top-$p$ to $0.95$, top-$k$ to $20$, repetition penalty to $1.0$, presence penalty to $1.5$, and the maximum number of generated tokens to $32768$.  To avoid prematurely truncating agent actions, we use the maximum context length supported by each model, namely 128K tokens for GLM-4.5 and 256K tokens for Qwen3.5, and allow each agent to execute up to 200 action steps. We run the experiments twice randomly and take the average.

\subsection{Details of Evaluation Benchmarks}
\label{subapp:benchmark_details}
\begin{itemize}
\item \textbf{BrowseComp-Plus}~\citep{BrowseComp-Plus}:
BrowseComp-Plus shares the same tasks as BrowseComp~\citep{BrowseComp}, but implements it through retrieval over a fixed local document corpus instead of live web search, enabling more reproducible comparisons across agent frameworks. Following the original benchmark setting, we use Qwen3-Embedding-8B~\citep{Qwen3-Embedding} as the retrieval model. Its key metric is answer-equivalence accuracy, which measures whether the predicted answer is semantically equivalent to the reference answer. In our experiments, we randomly sample 200 tasks for evaluation.

\item \textbf{WideSearch}~\citep{WideSearch}:
WideSearch focuses on broad information collection, where a system needs to identify a set of relevant entities and fill their attributes into a structured table. Its main metrics include Item-F1, Row-F1, and Success Rate: Item-F1 measures cell-level correctness, Row-F1 measures whether complete entity records are recovered, and Success Rate requires the whole table to match the reference answer.
The benchmark contains both Chinese and English tasks; in our experiments, we evaluate only the English subset, consisting of 100 tasks, to reduce resource costs.

\item \textbf{DeepWideSearch}~\citep{DeepWideSearch}:
DeepWideSearch covers tasks that require both deep search and broad information coverage. It is constructed in two ways: extending deep-search tasks into broad table-based information collection, and augmenting broad-search tasks with prerequisite deep-search steps. Its evaluation includes intermediate metrics, namely Column-F1 and Core Entity Accuracy, as well as final table-level metrics, including Item-F1, Row-F1, and Success Rate. Since the intermediate metrics in this benchmark are highly unreliable, we only report the final table-level metrics. The dataset contains both Chinese and English tasks; in our experiments, we evaluate only the English subset, consisting of 76 tasks, to reduce resource costs.

\item \textbf{GISA}~\citep{GISA}:
GISA is a general information-seeking benchmark covering multiple answer formats, including item, set, list, and table. Each format has corresponding metrics: Exact Match (EM) evaluates exact answer correctness, Set-F1 measures unordered set coverage, List-Content-F1 and List-Order-Score evaluate list completeness and ordering, and Table-Item-F1 and Table-Row-F1 evaluate structured table answers at the cell and row levels.
This benchmark is entirely in English; in our experiments, we evaluate all 373 tasks.
\end{itemize}

\subsection{Details of Baselines}
\label{subapp:baseline_details}

\begin{itemize} 
\item \textbf{ReAct Agent}~\citep{ReAct}: The ReAct Agent uses one agent to complete the entire task through interleaved reasoning and tool actions. It performs search, page reading, evidence analysis, and answer generation within a single context, without explicitly creating subagents or maintaining a structured task decomposition.

\item \textbf{Swarm-Agent}~\citep{Kimi-K2.5}: Swarm-Agent follows the inference framework of Kimi-Swarm~\citep{Kimi-K2.5}, where a coordinator can dynamically create subagents and assign subtasks to them. Concretely, the coordinator uses tools including \texttt{create\_subagents} to instantiate role-specific agents and \texttt{assign\_tasks} to delegate subtasks; the subagents then return their intermediate results to the coordinator for further reasoning, aggregation, or additional decomposition.

\item \textbf{Flash-Searcher}~\citep{Flash-Searcher}: Flash-Searcher organizes the search process as a dependency-aware directed acyclic graph (DAG) rather than a single linear trajectory. It decomposes a complex query into subtasks with explicit dependencies, tracks the logical constraints among them, and executes independent reasoning branches in parallel. The workflow can be dynamically optimized based on intermediate results, enabling more efficient tool use and reducing redundant sequential search steps.

\item \textbf{Table-as-Search}~\citep{Table-as-Search}: Table-as-Search represents long-horizon information seeking as a table-completion process. The external table acts as an explicit search state: rows usually correspond to candidate entities, columns correspond to constraints or target attributes, filled cells store verified information, and empty cells indicate remaining search targets.

\item \textbf{InfoSeeker}~\citep{InfoSeeker}: InfoSeeker adopts a hierarchical Host--Manager--Worker architecture. The Host performs high-level planning, Managers decompose and coordinate domain-specific subtasks, and Workers execute atomic subtasks. For a fair comparison, we replace the search APIs used in the original implementation with the same Serper-based search tool and Jina-based page-reading tool used in our experiments, while keeping the tool configuration consistent with other methods.

\item \textbf{ROMA}~\citep{ROMA}: ROMA is a recursive meta-agent framework that decomposes complex tasks into smaller subtasks and aggregates their results hierarchically. Its workflow typically consists of atomizing a task, planning dependency-aware subtasks, executing atomic subtasks, and aggregating the intermediate outputs through recursive control.
\end{itemize}

\section{Case Study}
\label{app:more_experiments}

To better illustrate how WebSwarm works across different complex search tasks, we provide several representative case studies as follows:
\begin{itemize}
\item \textbf{Figure~\ref{fig:case_study_bc_en_153_plain_tree} (BrowseComp-Plus).} 
This BrowseComp-Plus case requires identifying an unknown entity from multiple indirect constraints. WebSwarm instantiates a \textit{deep} search node, where multiple Searchers explore different clue paths and a Verifier checks the candidate against all constraints. The system finally identifies Walter Allen, demonstrating the candidate generation, verification, and refinement process for deep search.

\item \textbf{Figure~\ref{fig:case_study_ws_en_002_compact} (WideSearch).}
This case asks the system to build a product table across multiple spirits brands. Since the information is organized by brand on the web, WebSwarm expands into brand-level child nodes, first using \textit{entity\_collect} to obtain each brand's product set and then delegating \textit{atom} search nodes to fill product attributes. This illustrates recursive wide search following the structure ``brand$\rightarrow$product$\rightarrow$attribute.''

\item \textbf{Figure~\ref{fig:case_study_ws_taylor_swift_explore_atom} (WideSearch).}
This case requires extracting Taylor Swift's official concert records over a long time range. web-structure probing indicates that the data is mainly organized by tour rather than by year, so WebSwarm expands along official tours and delegates parallel \textit{atom} search nodes to extract show-level rows. This shows how web-structure-guided expansion helps choose a more effective expansion axis.

\item \textbf{Figure~\ref{fig:case_study_deep_wide_ford} (DeepWideSearch).}
This case first requires resolving a hidden manufacturer clue and then collecting structured vehicle information. WebSwarm first uses a \textit{deep} search node to identify Henry Ford, then expands into a \textit{wide} node that enumerates eligible Ford models with \textit{entity\_collect} and fills attributes with \textit{atom} nodes. This demonstrates an evidence-driven transition from deep entity identification to wide structured collection.
\end{itemize}

% ---------- Figure ----------
% =======================================================
% deepsearch figure
% =======================================================
    \begin{figure*}[t]
    \centering
    \begin{minipage}{0.96\textwidth}
    
    \centering
    % {\LARGE\bfseries\textcolor{tasnavy}{Case Study: bc\_plus\_92}}\par
    % \vspace{0.15em}
    
    % ---------- Original Question ----------
    \begin{tcolorbox}[
        enhanced,
        colback=white,
        colframe=tasnavy,
        boxrule=0.8pt,
        arc=2mm,
        title={Original Question},
        colbacktitle=tasnavy,
        coltitle=white,
        fonttitle=\bfseries\large,
        halign title=center,
        attach boxed title to top center={yshift=-1.8mm},
        boxed title style={
            colback=tasnavy,
            colframe=tasnavy,
            arc=1mm,
            left=8mm,
            right=8mm
        }
    ]
    \small
    Two people owned a business in a suburban town in the northwestern United States after 1900.
    The primary activity of the business focused on people and the surrounding area.
    The business moved locations to a street named for a number between 1 and 10.
    After 1910 but before 1920 one of the owners was no longer listed.
    The remaining owner worked with his wife and the business changed locations two more times.
    The wife's initials were G.F.P.
    What was the name of the owner who was no longer listed?
    \end{tcolorbox}
    
    % \vspace{0.2em}
    
    % ---------- One big structural box ----------
    \begin{tcolorbox}[
        enhanced,
        colback=white,
        colframe=gray!75!black,
        boxrule=0.8pt,
        arc=2mm,
        title={Recursive Deep Search Delegation},
        colbacktitle=gray!75!black,
        coltitle=white,
        fonttitle=\bfseries\large,
        halign title=center,
        left=2mm,
        right=2mm,
        top=1.5mm,
        bottom=1.5mm
    ]
    
    \small
    \noindent
    
    % ---------- Root: Orchestrator ----------
    \rootnode{tasred}{0.98\linewidth}{
    \textbf{\textcolor{tasred!85!black}{Root Node}}\\[-0.1em]
    {\footnotesize Receive the original question and initiate recursive delegation.}
    }
    
    % ---------- Level 1: Assigned subtask ----------
    \treenode{1.3em}{gray}{0.87\linewidth}
    {Delegate Subtask 1}
    {Find a business owned by two people in a suburban town in the northwestern US after 1900. The business focused on people and the surrounding area (likely a service business). It moved to a street named for a number between 1 and 10. After 1910 but before 1920, one of the original owners was no longer listed as an owner. The remaining owner continued working with his wife, whose initials were G.F.P. The business changed locations two more times. What was the name of the owner who was no longer listed?}
    
    % ---------- Level 1: Deep ----------
    \treenode{1.3em}{tasblue}{0.87\linewidth}
    {Deep Node}
    {Receive \textbf{Subtask 1}, organize iterative searcher--verifier through child search nodes.}
    
    % ---------- Level 2: Searchers and verifier under Deep ----------
    \treenode{3.8em}{tasblue}{0.78\linewidth}
    {Searcher 1}
    {Task: two people northwestern US numbered street 1910--1920; wife initials G.F.P.\\
    Output: Unknown.}
    
    \treenode{3.8em}{tasblue}{0.78\linewidth}
    {Searcher 2}
    {Task: wife initials G.F.P.; northwestern US 1910s service company partnership.\\
    Output: Unknown.}
    
    \treenode{3.8em}{tasblue}{0.78\linewidth}
    {Searcher 3}
    {Task: business moved First/Second/Third Street; partnership dissolved 1910--1920; spouse.\\
    Output: Unknown.}
    
    \treenode{3.8em}{tasblue}{0.78\linewidth}
    {Searcher 4}
    {Task: photography studio partnership dissolved 1910--1920; northwestern United States; wife on numbered street.\\
    Output: {Found candidate: the photography studio partnership of Frank Perkins and Walter Allen was dissolved around 1916-1917...}}
    
    \treenode{3.8em}{tasblue}{0.78\linewidth}
    {Searcher 5}
    {Task: Frank Perkins Walter Allen Georgetown Photograph Studio moved locations after 1916.\\
    Output: {Found supporting move history. (Frank Perkins Walter Allen's Georgetown Photograph Studio in Seattle moved locations multiple times after 1916...)}}
    
    \treenode{3.8em}{taspurple}{0.78\linewidth}
    {Verifier}
    {Check whether Walter Allen satisfies all constraints: a two-owner business after 1900; relocation to a numbered street; one owner no longer listed between 1910 and 1920...\\
    Output: \textcolor{tasgreen}{\cmark\ Verified: all constraints match.}}
    
    % ---------- Level 2: Deep synthesis ----------
    \treenode{1.3em}{tasblue}{0.87\linewidth}
    {Deep Node Result}
    {\textbf{Walter Allen} \textcolor{tasgreen}{\cmark}}
    
    % ---------- Orchestrator completion ----------
    \rootnode{tasred}{0.98\linewidth}
    {\textbf{\textcolor{tasred!85!black}{Root Node completion}}\\[-0.1em]
    {Receive the returned result and determine whether the original task is solved.
\textbf{Answer obtained. Task complete.} \textcolor{tasgreen}{\cmark}}
    }
    
    \vspace{0.7em}
    
    \end{tcolorbox}
    
    \end{minipage}
    
    \caption{Case study for bc\_plus\_92 in the BrowseComp-Plus benchmark.}
    \label{fig:case_study_bc_en_153_plain_tree}
    \end{figure*}
    
% =======================================================
% widesearch figure
% =======================================================
\begin{figure*}[t]
\centering
\begin{minipage}{0.96\textwidth}

\centering

% ---------- Original Question ----------
\begin{tcolorbox}[
    enhanced,
    colback=white,
    colframe=tasnavy,
    boxrule=0.75pt,
    arc=1.8mm,
    title={Original Question},
    colbacktitle=tasnavy,
    coltitle=white,
    fonttitle=\bfseries\large,
    halign title=center,
    attach boxed title to top center={yshift=-1.6mm},
    boxed title style={
        colback=tasnavy,
        colframe=tasnavy,
        arc=1mm,
        left=7mm,
        right=7mm,
        top=0.5mm,
        bottom=0.5mm
    },
    left=1.5mm,
    right=1.5mm,
    top=1.5mm,
    bottom=1mm
]
\small
I'm currently mapping the product portfolios of several spirits brands including
Johnnie Walker, Chivas Regal, Smirnoff, Grey Goose, Absolut Vodka, and Bacardi as of June 2025.
The scope covers only their standard products in the Core / Permanent Range, excluding any flavor variants
or limited-edition / seasonal releases.
Please organize the results in one Markdown table with the columns:
Brand, Product, Category, Sub-category, Pack Size (Bottle), ABV \%.
For any missing information, fill in `/`.
Please ensure that no cells in the Brand column are left blank.
Do not ask questions; directly output the results in the required table format.
\end{tcolorbox}

% ---------- One big structural box ----------
\begin{tcolorbox}[
    enhanced,
    colback=white,
    colframe=gray!75!black,
    boxrule=0.75pt,
    arc=1.8mm,
    title={Recursive Wide Search Delegation},
    colbacktitle=gray!75!black,
    coltitle=white,
    fonttitle=\bfseries\large,
    halign title=center,
    left=1.8mm,
    right=1.8mm,
    top=1.2mm,
    bottom=1.2mm
]

% ---------- Root ----------
\compactroot{tasred}{1.0\linewidth}
{Root Node}
{receive the request, initialize recursive delegation, and create a top-level \textit{wide} search node.}

% ---------- Level 1 ----------
\compactnode{1.1em}{tasblue}{0.88\linewidth}
{wide Search Node (depth=0)}
{expand along the brand dimension and delegate six brand-level child nodes.}

% ---------- Explore-guided parallel group ----------
\par\vspace{0.25em}
\noindent
\hspace*{2.8em}
\elbowarrow{tasblue}
\hspace{0.28em}
\begin{minipage}{0.81\linewidth}
\begin{tcolorbox}[
    enhanced,
    colback=tasblue!1,
    colframe=tasblue!60!black,
    boxrule=0.5pt,
    arc=1mm,
    frame style={dashed},
    title={\small Web-probing-guided expansion: distributed by brand},
    colbacktitle=tasblue!8,
    coltitle=tasblue!85!black,
    fonttitle=\bfseries,
    left=1.2mm,
    right=1.2mm,
    top=1mm,
    bottom=0.8mm
]

{\footnotesize
\textbf{\textcolor{taspurple!85!black}{web-probing agent:}}
evidence distribution = \textbf{distributed}; organization dimension = \textbf{brand}.
No single hub page covers all brands, so six brand-level child nodes are delegated in parallel.
}

\vspace{0.2em}

\begin{minipage}[t]{0.49\linewidth}
\brandwideexpanded{Johnnie Walker}{1}{8}{
\atomagent{1}{Red Label}
\atomagent{2}{Black Label}
\exptransfer{the official Johnnie Walker product pages consistently provide the needed fields, including product name...}
\atomagent{3}{Double Black Label}
\\ \quad \quad  ...
}{
Red Label, Black Label, ...
}
\brandwideexpanded{Grey Goose}{4}{3}{
\atomagent{1}{GREY GOOSE Vodka}
\atomagent{2}{GREY GOOSE Altius}
\\ \quad \quad ...
}{GREY GOOSE Vodka, Altius, ...}
\brandwideexpanded{Smirnoff}{3}{3}{
\atomagent{1}{Smirnoff White}
\atomagent{2}{No.21 Vodka}
\\ \quad \quad ...
}{Smirnoff White, No.21 Vodka, ...}
\end{minipage}
\hfill
\begin{minipage}[t]{0.49\linewidth}
\brandwideexpanded{Chivas Regal}{2}{5}{
\atomagent{1}{Chivas Regal XV}
\atomagent{2}{Chivas Regal Extra}
\exptransfer{search with the template ``Chivas Regal [Product Name] product details specifications'' to quickly locate information...}
\atomagent{3}{Chivas Regal Ultis}
\\ \quad \quad ...
}{Chivas Regal XV, Chivas Regal Extra, ...}
\brandwideexpanded{Absolut Vodka}{5}{3}{
\atomagent{1}{Absolut Elyx}
\atomagent{2}{Absolut Vodka}
\\ \quad \quad ...
}{Absolut Elyx, Absolut Vodka, ...}
\brandwideexpanded{Bacardi}{6}{7}{
\atomagent{1}{Bacardi Gold}
\atomagent{2}{Bacardi Superior}
\\ \quad \quad ...
}{Bacardi Gold, Bacardi Superior, ...}
\end{minipage}
\vspace{-1.55em}
\end{tcolorbox}
\end{minipage}
\par

\compactnode{1.1em}{tasblue}{0.88\linewidth}
{Wide Node Result}
{return the merged Markdown table upward.}

% ---------- Completion ----------
\compactroot{tasred}{1.0\linewidth}
{Root Node completion}
{receive the merged table and terminate the task.
% \medskip
\begin{center}
\scriptsize
% \resizebox{\linewidth}{!}{%
\begin{tabular}{llllll}
\toprule
Brand & Product & Category & Sub-category & Pack Size (Bottle) & ABV \% \\
\midrule
Johnnie Walker & Red Label & Whisky & Blended Scotch Whisky &
750ml & 40\% \\
... & ...& ...& ...& ...& ...\\
\bottomrule
\end{tabular}%
% }
\end{center}
}

\end{tcolorbox}

\end{minipage}

\caption{Case study for WS\_EN\_002 in the WideSearch benchmark.}
\label{fig:case_study_ws_en_002_compact}
\end{figure*}

% =======================================================
% explore figure
% =======================================================
\begin{figure*}[t]
\centering
\begin{minipage}{0.96\textwidth}

\centering

% ---------- Original Question ----------
\begin{tcolorbox}[
    enhanced,
    colback=white,
    colframe=tasnavy,
    boxrule=0.75pt,
    arc=1.8mm,
    title={Original Question},
    colbacktitle=tasnavy,
    coltitle=white,
    fonttitle=\bfseries\large,
    halign title=center,
    attach boxed title to top center={yshift=-1.6mm},
    boxed title style={
        colback=tasnavy,
        colframe=tasnavy,
        arc=1mm,
        left=7mm,
        right=7mm,
        top=0.5mm,
        bottom=0.5mm
    },
    left=1.5mm,
    right=1.5mm,
    top=1.5mm,
    bottom=1mm
]
\small
Could you list every single concert on Taylor Swift's official tour from January 1, 2010, to May 1, 2025,
including the specific date, the concert's English name, the country, the city, and the venue.
Each show should be on its own line, in chronological order from earliest to latest.

Please organize the results in one Markdown table with the following columns:
Date, the Concert's English Name, Host Country, Host City, Host Venue.
Do not use date ranges for Date; list it in the format of ``Day Month, Year'', for example, 4th June, 2011.
Do not ask follow-up questions; directly output the results according to the required columns.
\end{tcolorbox}

% ---------- One big structural box ----------
\begin{tcolorbox}[
    enhanced,
    colback=white,
    colframe=gray!75!black,
    boxrule=0.75pt,
    arc=1.8mm,
    title={Recursive Wide Search Delegation},
    colbacktitle=gray!75!black,
    coltitle=white,
    fonttitle=\bfseries\large,
    halign title=center,
    left=1.8mm,
    right=1.8mm,
    top=1.2mm,
    bottom=1.2mm
]

% ---------- Root ----------
\compactroot{tasred}{1.0\linewidth}
{Root Node}
{receive the original task, create a wide node, and assign it a task.}

% ---------- wide explore ----------
\compactnode{1.1em}{tasblue}{0.88\linewidth}
{wide Node}
{List every single concert on Taylor Swift's official tours from Jan. 1, 2010 to May 1, 2025. 
For each show, extract the specific date, concert English name, host country, host city, and venue.
Sort all rows chronologically and output one Markdown table without omitted cells.}

% ---------- Explore agent ----------
\par\vspace{0.2em}
\noindent
\hspace*{2.8em}
\elbowarrow{taspurple}
\hspace{0.28em}
\begin{minipage}{0.81\linewidth}
\explorecontext{1.0\linewidth}
\end{minipage}
\par\vspace{-1.0em}

% ---------- Atom group ----------
\par\vspace{0.0em}
\noindent
\hspace*{2.8em}
\elbowarrow{tasorange}
\hspace{0.28em}
\begin{minipage}{0.81\linewidth}
\begin{tcolorbox}[
    enhanced,
    colback=tasorange!1,
    colframe=tasorange!65!black,
    boxrule=0.5pt,
    arc=1mm,
    % dashed,
    title={\small Six tour-level atom agents dispatched by wide and executed in parallel},
    colbacktitle=tasorange!8,
    coltitle=tasorange!90!black,
    fonttitle=\bfseries,
    left=1.2mm,
    right=1.2mm,
    top=1mm,
    bottom=0.4mm,
    before skip=0pt,
    after skip=0pt
]

\begin{minipage}[t]{0.49\linewidth}
\atomtour{Fearless Tour}{1/6}{Fearless\_Tour}
\atomtour{Speak Now World Tour}{3/6}{Speak\_Now\_World\_Tour}
\atomtour{The Red Tour}{5/6}{The\_Red\_Tour}
\end{minipage}
\hfill
\begin{minipage}[t]{0.49\linewidth}
\atomtour{The 1989 World Tour}{2/6}{The\_1989\_World\_Tour}
\atomtour{Reputation Stadium Tour}{4/6}{Reputation\_Stadium\_Tour}
\atomtour{The Eras Tour}{6/6}{The\_Eras\_Tour}
\end{minipage}
\vspace{-1.15em}
\end{tcolorbox}
\end{minipage}

\par

% ---------- Final output ----------
\compactnode{1.1em}{tasblue}{0.88\linewidth}
{Wide Node Result}
{aggregate child-node results into a table and return it upward.}

% ---------- Completion ----------
\compactroot{tasred}{1.0\linewidth}
{Root Node completion}
{receive the returned chronological table and terminate the task.
\begin{center}
\scriptsize
% \resizebox{\linewidth}{!}{%
\begin{tabular}{lllll}
\toprule
Date & Concert Name & Host Country & Host City & Host Venue\\
\midrule
6th February, 2010  & Fearless Tour & Australia & Sydney & Acer Arena \\
7th February, 2010  & Fearless Tour & Australia & Sydney & Acer Arena\\
...& ...& ...& ...& ...\\
\bottomrule
\end{tabular}%
% }
\end{center}
}

\end{tcolorbox}

\end{minipage}

\caption{Case study for WS\_EN\_006 in the WideSearch benchmark.}
\label{fig:case_study_ws_taylor_swift_explore_atom}
\end{figure*}

% =======================================================
% deep wide
% =======================================================
\begin{figure*}[t]
\centering
\begin{minipage}{0.96\textwidth}

\centering

% ---------- Original Question ----------
\begin{tcolorbox}[
    enhanced,
    colback=white,
    colframe=tasnavy,
    boxrule=0.75pt,
    arc=1.8mm,
    title={Original Question},
    colbacktitle=tasnavy,
    coltitle=white,
    fonttitle=\bfseries\large,
    halign title=center,
    attach boxed title to top center={yshift=-1.6mm},
    boxed title style={
        colback=tasnavy,
        colframe=tasnavy,
        arc=1mm,
        left=7mm,
        right=7mm,
        top=0.5mm,
        bottom=0.5mm
    },
    left=1.5mm,
    right=1.5mm,
    top=1.5mm,
    bottom=1mm
]
\small
Identify all US vehicles first launched or resumed production between 2010 and 2024,
where the manufacturer is associated with a US entrepreneur who built experimental mobile devices,
later pioneered continuous-flow industrial production methods, and whose family-owned business
retains control through special voting rights. For each standard launch edition, output launch MSRP,
dimensions, wheelbase, torque, suspension, ADAS, and IPA in one Markdown table.
\end{tcolorbox}

% ---------- One big structural box ----------
\begin{tcolorbox}[
    enhanced,
    colback=white,
    colframe=gray!75!black,
    boxrule=0.75pt,
    arc=1.8mm,
    title={Recursive Deep-Wide Search Delegation},
    colbacktitle=gray!75!black,
    coltitle=white,
    fonttitle=\bfseries\large,
    halign title=center,
    left=1.8mm,
    right=1.8mm,
    top=1.2mm,
    bottom=1.2mm
]

% ---------- Root ----------
\compactroot{tasred}{1.0\linewidth}
{Root Node}
{first delegate hidden-entity identification, and then collect vehicle table based on the returned evidence.}

% ---------- Deep phase ----------
\compactnode{1.1em}{taspurple}{0.88\linewidth}
{Deep Node}
{{Identify the entrepreneur and associated manufacturer implied by the hidden clues: experimental vehicles, continuous-flow industrial production, and family control through special voting rights.}}

% ---------- Deep subagents listed directly, without an enclosing group box ----------
\compactnode{3.8em}{taspurple}{0.78\linewidth}
{searcher 1}
{Task: US entrepreneur vehicle manufacturing continuous-flow industrial production methods / moving assembly line pioneer.\\
Output: \textcolor{tasgreen}{\cmark\ Candidate identified: Henry Ford.}}

\compactnode{3.8em}{taspurple}{0.78\linewidth}
{verifier}
{Task: verify Henry Ford's experimental mobile devices before the Model T, including the 1896 Quadricycle, Sweepstakes racer, and pre-Model-T letter cars.\\
Output: \textcolor{tasgreen}{\cmark\ Experimental-vehicle clue matches Henry Ford.}}

\compactnode{3.8em}{taspurple}{0.78\linewidth}
{verifier}
{Task: verify Ford Motor Company family control through special voting rights / Class B shares.\\
Output: \textcolor{tasgreen}{\cmark\ Family-control clue matches Ford.}}

% ---------- Bridge ----------
\compactnode{1.1em}{taspurple}{0.88\linewidth}
{Deep output}
{\textbf{Henry Ford}}

% ---------- Wide phase ----------
\compactnode{1.1em}{tasblue}{0.88\linewidth}
{wide(depth=0)}
{collect all eligible Ford models and dispatch one atom agent per vehicle to fill the requested columns.}

\par\vspace{0.25em}
\noindent
\hspace*{2.8em}
\elbowarrow{tasblue}
\hspace{0.28em}
\begin{minipage}{0.81\linewidth}

\dwvehiclewideexpanded{Ford vehicles}{1/1}{16}{
\begin{minipage}[t]{0.48\linewidth}
\dwvehicleatom{1}{Focus Electric (2012)}
\dwvehicleatom{2}{C-Max Energi (2013)}
\dwvehicleatom{3}{Fusion Energi (2013)}
\par\vspace{0.05em}
\hspace*{1.15em}{\scriptsize\textbf{...}}
\end{minipage}
\hfill
\begin{minipage}[t]{0.48\linewidth}
\dwvehicleatom{4}{EcoSport (2018)}
\dwvehicleatom{5}{GT (2017)}
\dwvehicleatom{6}{Ranger (2019)}
\end{minipage}
}{
Focus Electric, C-Max Energi, Fusion Energi, EcoSport, GT, Ranger, Bronco,
Mustang Mach-E, Maverick, F-150 Lightning, Bronco Sport, Transit Connect,
C-Max Hybrid, F-150 Raptor, Escape Plug-in Hybrid, E-Transit
}

\end{minipage}
\par
\vspace{-1.0em}
\compactnode{1.1em}{tasblue}{0.88\linewidth}
{Wide output}
{generate one Markdown table.
}

% ---------- Completion ----------
\compactroot{tasred}{1.0\linewidth}
{Root Node Completion}
{receive the returned Markdown table from the Wide node and terminate the task.\begin{center}
\scriptsize
\resizebox{\linewidth}{!}{%
\begin{tabular}{lllllllll}
\toprule
Model Name & Price & Dimensions L/W/H & Wheelbase & Torque & Front Susp. & Rear Susp. & ADAS & IPA \\
\midrule
Focus Electric (2012) & \$39,200 & 4392/1823/1478 & 2649 & 245 & MacPherson & Multi-Link & Rear Parking Sensor, ABS, ... & - \\
C-Max Energi (2013) & \$32,950 & 4409/1828/1621 & 2648 & 239 & MacPherson & Multi-Link & AdvanceTrac, traction control, ... & - \\
... & ... & ... & ... & ... & ... & ... & ... & ... \\
\bottomrule
\end{tabular}%
}
\end{center}}

\end{tcolorbox}

\end{minipage}

\caption{Case study for Wide2Deep\_ws\_en\_049 in the DeepWideSearch benchmark.}
\label{fig:case_study_deep_wide_ford}
\end{figure*}

\begin{figure*}[p]
\centering
\begin{promptbox}{Root Agent System Prompt}
You are a **research orchestrator**. You receive a research task from the user and resolve it by dispatching subtasks to specialized verb agents.

Current Date: {current_date}

## How You Operate
You are a **single-layer planner**: each turn you must call exactly ONE tool - either `solve_subtask` (to delegate a subtask) or `submit_answer` (to finalize). After every `solve_subtask` call you will see the result, then decide the next action.

You do NOT execute web searches yourself. All retrieval and reasoning happens inside verb agents.

## Available Verbs
You must classify each subtask into exactly one of:

### `atom`
- Single-entity attribute lookup or short multi-hop chain.
- Use when the entity is named (or is the unique result of a clear chain) and the answer is a small set of attributes.
### `deep`
- The target entity is unknown and must be uncovered from a combination of indirect / vague constraints.
- Hypothesize candidates -> verify against constraints -> narrow down. A single keyword search will not find it.
### `wide`
- Same kind of information collected over a group of items (fan-out / table-fill).
- The task has at least one iteration dimension (years, brands, countries, ...) and a per-item attribute set.
- Use this whether the iteration list is already given OR needs internal discovery - the wide agent handles both.
### `entity_collect`
- Enumerate a complete entity set with high precision and high recall.
- The output IS the set itself; per-item attributes (if any) are secondary.

## Decision Heuristics
- One named entity + a few attributes -> `atom`.
- Many items * per-item attributes -> `wide`.
- Output is a set / list whose membership itself is the question -> `entity_collect`. (If the task also asks for per-item attributes alongside a list of unknown items, prefer `wide` - wide agents will discover the list internally.)
- Constraint-intersection puzzle, target entity unnamed -> `deep`.

## Workflow
1. **Read the task** and decide whether ONE verb call can cover the whole thing, or whether the task naturally splits into a small sequence of dependent subtasks (each subtask itself goes to a single verb).
2. **Dispatch** via `solve_subtask(task, verb)`. The `task` field must be **fully self-contained** - include all dates, scopes, exclusions, and the requested output format, since the verb agent has no access to the original user message or your prior reasoning.
3. **Read the result** carefully. If it is incomplete, inconsistent, or fails to answer the original question, dispatch a follow-up subtask (often a refined `atom` or another verb).
4. **Submit** via `submit_answer(answer)` once the original user task is fully answered.

## Rules
- Mandatory tool use: call a tool every turn.
- One tool call per turn. Do NOT batch multiple verbs into one turn.
- Each subtask must be self-contained.
- The `verb` field of `solve_subtask` MUST be one of: `atom`, `deep`, `wide`, `entity_collect`.
- Do NOT rely on internal knowledge - only on verb agents' returned results.
- **Zero-assumption principle**: When the task references categories, lists, rankings, or classifications from a named source, you MUST NOT fill in those items from your own knowledge. Instead, treat them as **unknown** and let the verb agent discover them by searching the authoritative source.
- If the user asks for a table, the final `submit_answer` must format the answer as ```markdown\n{table content}\n```.
\end{promptbox}
\caption{System prompt for guiding the root agent to create and dispatch subtasks to search nodes.}
\label{fig:root-agent-prompt}
\end{figure*}

\begin{figure*}[p]
\centering
\begin{promptbox}{Web-Probing Agent Prompt}
You are an **information topology scout**. Before a research task is decomposed into parallel sub-tasks, you do a quick search to determine how the required information is distributed across web pages. Current Date: {current_date}

## Your Goal
Determine the **information topology** for a research task:
1. **centralized** - 1-3 hub pages cover ALL required data (rows AND columns).
2. **centralized_with_gaps** - hub pages cover a large portion of the required data, but have identifiable gaps (missing columns, incomplete rows, missing details, etc.).
3. **distributed** - no hub covers even the majority of the data; information is scattered across many independent pages.

## Workflow
1. Read the task. Identify the key entities (rows) and attributes (columns) needed.
2. Search for aggregate / hub sources - **prioritize Wikipedia**:
   - Wikipedia "List of ..." pages are often the best structured hub sources. Try queries like `<topic> site:en.wikipedia.org` or `List of <entities>` first.
   - Also try: "[topic] complete list", "[topic] table / database / all"
3. Fetch promising pages and **check coverage column by column**:
   - For each required column, confirm whether the page actually contains it (not just whether a table exists).
   - Check row scope: does the page cover the full range the task asks for (time range, geographic scope, etc.)?
   - A page having a table does NOT mean it covers everything - verify.
(Repeat steps 2 and 3 until a clear conclusion is reached.)
4. Make your topology judgment and submit.

## Output Format - pick ONE of the three topologies:

### If data is fully centralized (1-3 pages cover ALL rows AND columns):

## Topology
centralized
## Hub Sources
- <URL 1> - <what rows/columns this page covers>
## Page Structure
<brief factual description: single table / multiple tables / paginated / etc.>

### If data is centralized with gaps (hub covers most data, but has identifiable gaps):

## Topology
centralized_with_gaps
## Hub Sources
- <URL 1> - <what this page covers>
## Hub Coverage
<what the hub page(s) DO provide: which rows, columns, or details are available>
## Gaps
<what is missing: specific columns not present, rows not covered, details absent, etc.>
## Estimated Coverage
<rough fraction of required data on hub, e.g. "4/5 columns", "~80% of rows">

### If data is distributed (no hub covers the majority of rows):

## Topology
distributed
## Organization Dimension
<the dimension along which pages are organized: e.g., "tour name", "brand", "program", "year", "country">
## Evidence
<factual observation about page structure that supports this dimension>
## Key Source Pattern
- <representative URL pattern or hub page> - <what it links to>
\end{promptbox}
\caption{The system prompt for guiding the agent to probe Web information structures.}
\label{fig:explore-agent-prompt}
\end{figure*}

\begin{figure*}[p]
\centering
\begin{promptbox}{Subtask Experience Extraction Prompt}
You are a skill-extraction assistant for a multi-agent research system.

## CONTEXT
A "wide" task has been decomposed into a batch of sibling sub-tasks (the FANOUT batch). The system first ran a small SCOUT batch (the first N sub-tasks) and recorded their full execution traces. Now,
**before** dispatching the remaining sub-tasks, you extract transferable tactical knowledge ("skill") from the scout traces to help the remaining sub-agents work more efficiently and more effectively.

## YOU WILL BE GIVEN
  - ROOT_TASK        : the original user-level question
  - WIDE_TASK        : the parent wide task that produced this batch
  - SCOUT_TASKS      : the sub-tasks that already completed (with answers)
  - REMAINING_TASKS  : the sub-tasks that have NOT yet been dispatched
  - SCOUT_TRACES     : full message traces of the scout sub-agents

## YOUR JOB
Analyze the scout execution traces and extract transferable process
skill such as:
  - which queries / domains worked well / poorly
  - which URLs were authoritative for this kind of question
  - which dead ends to skip
  - any constraint the scout discovered that affects all siblings

Do NOT include facts that are specific to a single scout sub-task's answer. The skill is process-level, not knowledge-level.

## HARD RULES
  1. Skill text must be **process advice**, not factual claims. It will be prepended to each sub-agent's task brief as if a sibling were whispering "here's what worked for me".
  2. Output ONLY the skill text itself - no JSON, no code fences, no preamble, no explanation. Just the plain-text skill paragraph.
  3. Keep it concise: <= 150 words, plain English, no markdown headers.
  4. If no useful tactical skill is observable from the scout traces, output exactly the single word: NONE
\end{promptbox}
\caption{The system prompt for guiding the LLM to summarize reusable subtask experience.}
\label{fig:skill-extraction-prompt}
\end{figure*}

\begin{figure*}[p]
\centering
\begin{promptbox}{Web Tool Interface}
{
    "name": "search",
    "description": "Search for relevant information on the web with a single query. Important note: Choose query language based on information source type.",
    "parameters": {
        "type": "object",
        "properties": {
            "query": {
                "type": "string",
                "description": "A single search query string. Language selection guide: choose the language that can obtain the most relevant results from the primary information sources."
            },
            "date_range": {
                "type": "string",
                "description": "Time range (optional), default any time.",
                "enum": ["qdr:h", "qdr:d", "qdr:w", "qdr:m", "qdr:y"]
            }
        },
        "required": ["query"]
    }
}

{
    "name": "fetch_url",
    "description": "Fetch content from a single URL, and extract the content you want by an intelligent agent.",
    "parameters": {
        "type": "object",
        "properties": {
            "url": {
                "type": "string",
                "description":  "A single URL to fetch."
            },
            "goal": {
                "type": "string",
                "description": "The goal of the content you want to extract from the web page."
            }
        },
        "required": ["url","goal"]
    }
}
\end{promptbox}
\caption{Descriptions of the Web-Search Tool and the Page-Browse Tool provided to agents.}
\label{fig:web-tool-interface}
\end{figure*}

\end{document}